%% file: main.tex
\documentclass[10pt,twocolumn,letterpaper]{article}

\usepackage{cvpr}              
\usepackage{multirow}
\usepackage{makecell}
\usepackage{bbding}
\usepackage{enumitem}
\usepackage{colortbl}
\usepackage{xcolor}
\usepackage{balance}
\usepackage{listings}
\definecolor{myblue}{RGB}{233, 241, 249}
\definecolor{mygray}{RGB}{99, 110, 114}
\definecolor{myred}{RGB}{255, 118, 117}
\definecolor{myyellow}{RGB}{255, 234, 167}
\definecolor{mygreen}{RGB}{216, 226, 204}

\include{}

\definecolor{cvprblue}{rgb}{0.21,0.49,0.74}
\usepackage[pagebackref,breaklinks,colorlinks,allcolors=cvprblue]{hyperref}

\def\paperID{3669} 
\def\confName{CVPR}
\def\confYear{2025}

\usepackage{CJKutf8}

\title{ECBench: Can Multi-modal Foundation Models Understand the Egocentric World?  A Holistic Embodied Cognition Benchmark}

\author{
Ronghao Dang\textsuperscript{1}\thanks{Equal contribution.} , Yuqian Yuan\textsuperscript{2}\footnotemark[1] , Wenqi Zhang\textsuperscript{2}\footnotemark[1] , Yifei Xin\textsuperscript{1}, Boqiang Zhang\textsuperscript{1}, \\
Long Li\textsuperscript{2}, Liuyi Wang\textsuperscript{3}, Qinyang Zeng\textsuperscript{3}, Xin Li\textsuperscript{1}\thanks{Corresponding author.}, Lidong Bing\textsuperscript{4} \\
{\small \textsuperscript{1}Alibaba DAMO Academy, \textsuperscript{2}Zhejiang University,
\textsuperscript{3}Tongji University,
\textsuperscript{4}Shanda AI Research Institute}\\
{\tt\small dangronghao.drh@alibaba-inc.com}
}

\begin{document}
\maketitle
\input{sec/0_abstract}    
\input{sec/1_intro}
\input{sec/2_related_works}
\input{sec/3_ECBench}

\input{sec/4_Experiment}
\input{sec/5_Conclusion}

\section*{Acknowledgments}
This paper is supported by the National Natural Science Foundation of China under Grants (624B2105).

{
    \small
    \bibliographystyle{ieeenat_fullname}
    \bibliography{main}
}

\input{sec/X_suppl}

\end{document}

%% file: sec/0_abstract.tex
\begin{abstract}
The enhancement of generalization in robots by large vision-language models (LVLMs) is increasingly evident. Therefore, the embodied cognitive abilities of LVLMs based on egocentric videos are of great interest. However, current datasets for embodied video question answering lack comprehensive and systematic evaluation frameworks. Critical embodied cognitive issues, such as robotic self-cognition, dynamic scene perception, and hallucination, are rarely addressed.
To tackle these challenges, we propose ECBench, a high-quality benchmark designed to systematically evaluate the embodied cognitive abilities of LVLMs. ECBench features a diverse range of scene video sources, open and varied question formats, and 30 dimensions of embodied cognition. To ensure quality, balance, and high visual dependence, ECBench uses class-independent meticulous human annotation and multi-round question screening strategies.
Additionally, we introduce ECEval, a comprehensive evaluation system that ensures the fairness and rationality of the indicators. Utilizing ECBench, we conduct extensive evaluations of proprietary, open-source, and task-specific LVLMs. ECBench is pivotal in advancing the embodied cognitive capabilities of LVLMs, laying a solid foundation for developing reliable core models for embodied agents. All data and code is available at https://github.com/Rh-Dang/ECBench.
\end{abstract}

%% file: sec/1_intro.tex
\section{Introduction}
\label{sec:intro}

\begin{figure*}[t]
  \centering
   \includegraphics[width=0.98\linewidth]{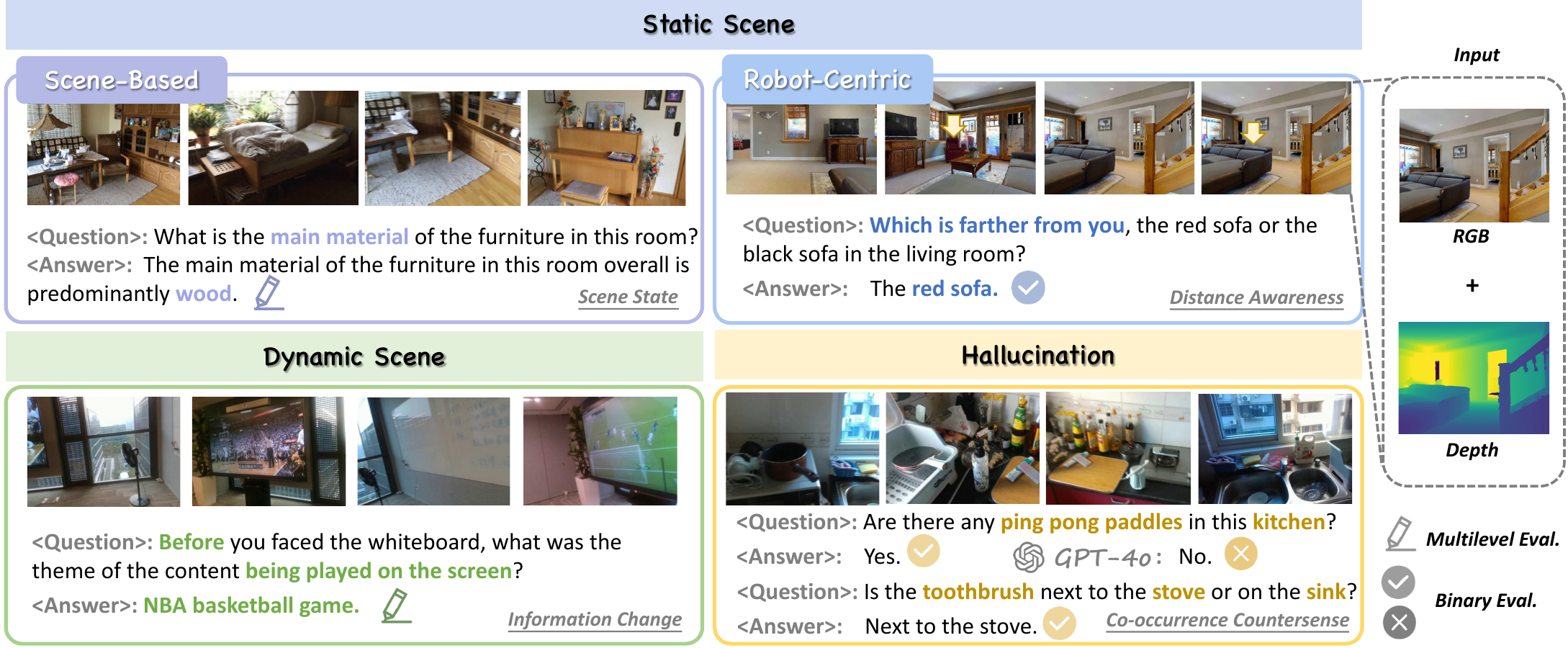}
   \caption{\textbf{Illustration of question answering (QA) format and representative cognitive dimensions from ECBench.} There are 386 RGB-D videos, 4,324 QA pairs, and 30 distinct embodied cognitive abilities, spanning across various aspects such as perception, reasoning, self-awareness, dynamic capturing, and hallucination. ECEval employs distinct evaluation methods for different types of answers.}
   \vspace{-5pt}
   \label{fig:introduction}
\end{figure*}


Spurred by the significant advancements in large language models (LLMs), AI researchers are now more optimistic than ever about achieving artificial general intelligence. Many pioneering studies aim to cognize the physical world by developing models capable of understanding multimodal inputs, such as MiniGPT-4 \cite{minigpt-4}, Video-LLaMA \cite{video-llama}, etc. 
However, these models are typically trained and evaluated on non-egocentric inputs. Therefore, it is crucial to examine the barriers to their application in embodied environments.

A comprehensive and reliable benchmark is essential for evaluating the multimodal understanding capabilities of models in egocentric embodied contexts.
Such benchmarks are very scarce due to the challenges involved in egocentric video collection and embodied questions annotation. ScanQA \cite{scanqa} and SQA3D \cite{sqa3d} are representative close-vocabulary benchmarks for 3D scene understanding. These task-specific datasets are insufficient to comprehensively assess task-generalized LVLMs. Therefore, OpenEQA \cite{openeqa} introduces the first open-vocabulary benchmark to assess foundational models' question-answering capabilities in indoor embodied videos. Despite these efforts, the current evaluation of LVLMs in embodied scenarios has the following limitations:

\begin{itemize}
    \item \textbf{Not Systematic:} Current benchmarks focus on independent embodied abilities, such as object recognition and counting. They lack a comprehensive top-down analysis of embodied cognition requirements, leading to limitations in evaluation hierarchy and dimensions.
    \item  \textbf{Lack of Robot-Centric:} As illustrated by the robot-centric example in Fig.~\ref{fig:introduction}, robots often need to address questions related to their own embodiment, such as the distance to a target, or their historical trajectory. However, benchmarks like OpenEQA focus solely on third-person scenario questions, significantly overlooking the evaluation of  robots' self-awareness.
    \item \textbf{Lack of Dynamics:} In the real physical world, scene dynamics are perpetually ongoing, as exemplified by video content changes on the screen in the dynamic scene example of Fig.~\ref{fig:introduction}.  For complex tasks like ``Revert the screen content to before you faced the whiteboard," a robot must recognize these dynamics and accurately recall their timing and process. However, current embodied question answering benchmarks typically overlook these dynamic aspects, defaulting to static context assumptions. 
    \item \textbf{Hallucination issue:} Although the hallucination phenomenon has been extensively analyzed within LVLMs, embodied-based question answering presents unique hallucination challenges. For instance, as depicted in the hallucination example of Fig.~\ref{fig:introduction}, LVLMs like GPT-4o often rely too much on common sense in the scene when answering questions in counterintuitive scenes, resulting in incorrect answers. These embodied hallucination issues remain unexplored in the academic literature.
\end{itemize}

\vspace{-0.2em}
To address the aforementioned challenges, we propose ECBench, a comprehensive video question answering benchmark designed to evaluate the embodied cognitive abilities of LVLMs. ECBench covers three sets: static scenes, dynamic scenes, and hallucinations. Within the static scene category, we systematically categorize questions into scene-based and robot-centric types, encompassing 19 cognitive abilities such as spatial reasoning and trajectory review.  Notably, we innovatively introduce robot-centric cognitive questions, aiming for the models to develop self-awareness and understand the relationship between their own entity and the environment. Robot-centric cognition is crucial when robots frequently move and perform complex tasks, such as searching and transporting.

Additionally, we propose the first open-world question answering task for dynamic scenes, encompassing dynamics both within and beyond immediate visibility. To highlight the model's capability in perceiving scene dynamics, all questions are related to changes within the scene. We focus on four categories:
spatial dynamics, information dynamics, quantity dynamics, and state dynamics.

We further evaluate hallucination issues specific to embodied scenes from seven detailed perspectives, grouped into two aspects: over-confidence in common sense and over-confidence in user input. Over-confidence in common sense impairs LVLMs' ability to handle counterintuitive scenarios; for instance, they might struggle to accept the visual fact of a toothbrush being next to a stove (Fig.~\ref{fig:introduction}). Over-confidence in user input hinders LVLMs' ability to recognize ambiguous, incorrect, or missing references in user input, substantially reducing the robot's interactivity during task execution.

\begin{table*}[t]
\centering
\small
\setlength\tabcolsep{2.8pt} 
\begin{tabular}{l|ccc|ccc|cccc}
\Xhline{2\arrayrulewidth}
\multirow{2}{*}{\textbf{Benchmark}} & \multicolumn{3}{c|}{\textbf{Visual Input}} & \multicolumn{3}{c|}{\textbf{Question Answering Pairs}} & \multicolumn{4}{c}{\textbf{Evaluation Domain}} \\
 & \begin{tabular}[c]{@{}c@{}}Real \\ Scenes\end{tabular} & RGB-D & Egocentric & \begin{tabular}[c]{@{}c@{}}Open \\ Vocab.\end{tabular} & Generation & QA Form & \begin{tabular}[c]{@{}c@{}}Number of \\ Capabilities\end{tabular} & \begin{tabular}[c]{@{}c@{}}Robot\\ Centric\end{tabular} & Halluc. & \begin{tabular}[c]{@{}c@{}}Dynamic \\ Scene\end{tabular} \\ \Xhline{2\arrayrulewidth}
ScanQA \cite{scanqa} & \Checkmark & \Checkmark & \Checkmark &  & Automatic/Human & Open & \textbf{—} &  &  &  \\
SQA3D \cite{sqa3d} & \Checkmark & \Checkmark & \Checkmark &  & Human & Open & \textbf{—} & \Checkmark &  &  \\
Env-QA \cite{env-qa} &  & \Checkmark & \Checkmark &  & Template & Open & 5 &  &  & \Checkmark \\
RoboVQA \cite{robovqa} & \Checkmark &  & \Checkmark & \Checkmark & Automatic & Open & 8 &  &  & \Checkmark \\
OpenEQA \cite{openeqa} & \Checkmark & \Checkmark & \Checkmark & \Checkmark & Human & Open & 7 &  &  &  \\ \Xhline{2\arrayrulewidth}
MMBench-Video \cite{mmbench-video} & \Checkmark &  &  & \Checkmark & Human & Open & 26 &  & \Checkmark & \Checkmark \\
MVBench \cite{mvbench} & \Checkmark &  &  & \Checkmark & Template/Automatic & Close & 20 &  &  & \Checkmark \\ \Xhline{2\arrayrulewidth}
\rowcolor{black!10}\textbf{ECBench(Ours)} & \Checkmark & \Checkmark & \Checkmark & \Checkmark & Human & Open/Close & 30 & \Checkmark & \Checkmark & \Checkmark \\ \Xhline{2\arrayrulewidth}
\end{tabular}
\caption{\textbf{Comparing ECBench and widely adopted Embodied / General VideoQA benchmarks.} ECBench has significant advantages in terms of quality, diversity and evaluation dimensions.}
\vspace{-7pt}
\label{tab:compare_benchmark}
\end{table*}

During dataset collection, to fulfill the evaluation requirements for different aspects of embodied cognition, we permit annotators to flexibly employ various question formats, including open-ended and multiple-choice questions. 
For evaluation, we propose ECEval, a novel assessment framework for question-answering tasks in embodied contexts. ECEval amalgamates binary scoring and multi-level scoring, enabling accurate evaluation of answer correctness for both open-ended and closed-ended questions (Fig.~\ref{fig:introduction}). Uniquely, we execute a multi-round screening strategy supported by GPT-4o blind testing to minimize the proportion of questions that can be answered solely based on common sense. Overall, the benchmark encompasses 4,324 unique question-answering (QA) pairs provided by volunteers, covering a total of 30 fine-grained evaluation angles.

Based on ECBench, we comprehensively evaluate the embodied cognitive abilities of various LVLMs. Our assessment encompasses open-source and proprietary general LVLMs \cite{qwen2Vl, gpt4o, kangaroo}, as well as embodied / egocentric LVLMs \cite{evud, multihop-egoqa}. 
Notably, all mainstream LVLMs exhibit poor performance in dynamic scenes and embodied hallucination issues. Furthermore, in static scenes, robot-centric questions (first-person) present greater challenges compared to scene-based questions (third-person). These results indicate that contemporary LVLMs only possess third-person cognition in static scenes, while they struggle to achieve first-person understanding in dynamic scenes. 
We aim for ECBench to drive the development of LVLMs toward more practical embodied scenarios.

%% file: sec/2_related_works.tex
\section{Related Work}
\label{sec:related_work}

We compare ECBench with popular general and embodied video question answering (VideoQA) benchmarks in Tab.~\ref{tab:compare_benchmark}.

\subsection{General VideoQA Benchmarks}
VideoQA technology promotes the model understanding of the world.
Initially, popular evaluation benchmarks concentrate on specific domains such as movies \cite{movieqa}, TV series \cite{tvqa,knowit}, and video games \cite{marioqa}.  However, with the advent of LVLMs, the scope of model applicability has broadened. To evaluate these LVLMs in all aspects, comprehensive benchmarks \cite{mmbench-video,mvbench,seed-bench,egoschema} have emerged. MMbench-video \cite{mmbench-video} uses a carefully constructed capability classification. MVBench \cite{mvbench} introduces a novel framework of spatial-temporal task construction. Nonetheless, general VideoQA benchmarks focus on YouTube videos and seldom incorporate embodied-specific QA formats. 
Our ECBench maintains the comprehensive and systematic nature of general benchmarks while far exceeding them in embodied cognition (e.g. 3D spatial understanding, robot self-awareness).

\subsection{Embodied VideoQA Benchmarks}
In embodied scenarios, VideoQA-based evaluations can effectively gauge a model's understanding of its environment and tasks. ScanQA \cite{scanqa}, SQA3D \cite{sqa3d}, and Env-QA \cite{env-qa}  are typical datasets for traditional scene question answering with a closed vocabulary. These datasets have a strong text bias and relatively limited question forms. RoboVQA \cite{robovqa}, EgoPlan-Bench \cite{egoplan-bench}, and PCA-Bench \cite{pca-bench} focus on testing task-planning abilities of LVLMs but lack direct evaluation of cognitive capabilities. OpenEQA \cite{openeqa} represents the first open vocabulary benchmark in the realm of embodied video understanding. However, it lacks robot-centric questions and dynamic scenes. ECBench is the first work to systematically analyze the embodied cognition of LVLMs, allowing researchers to explore the frontiers of LVLMs in robots.

%% file: sec/3_ECBench.tex
\section{ECBench}

\begin{figure*}[t]
  \centering
   \includegraphics[width=0.95\linewidth]{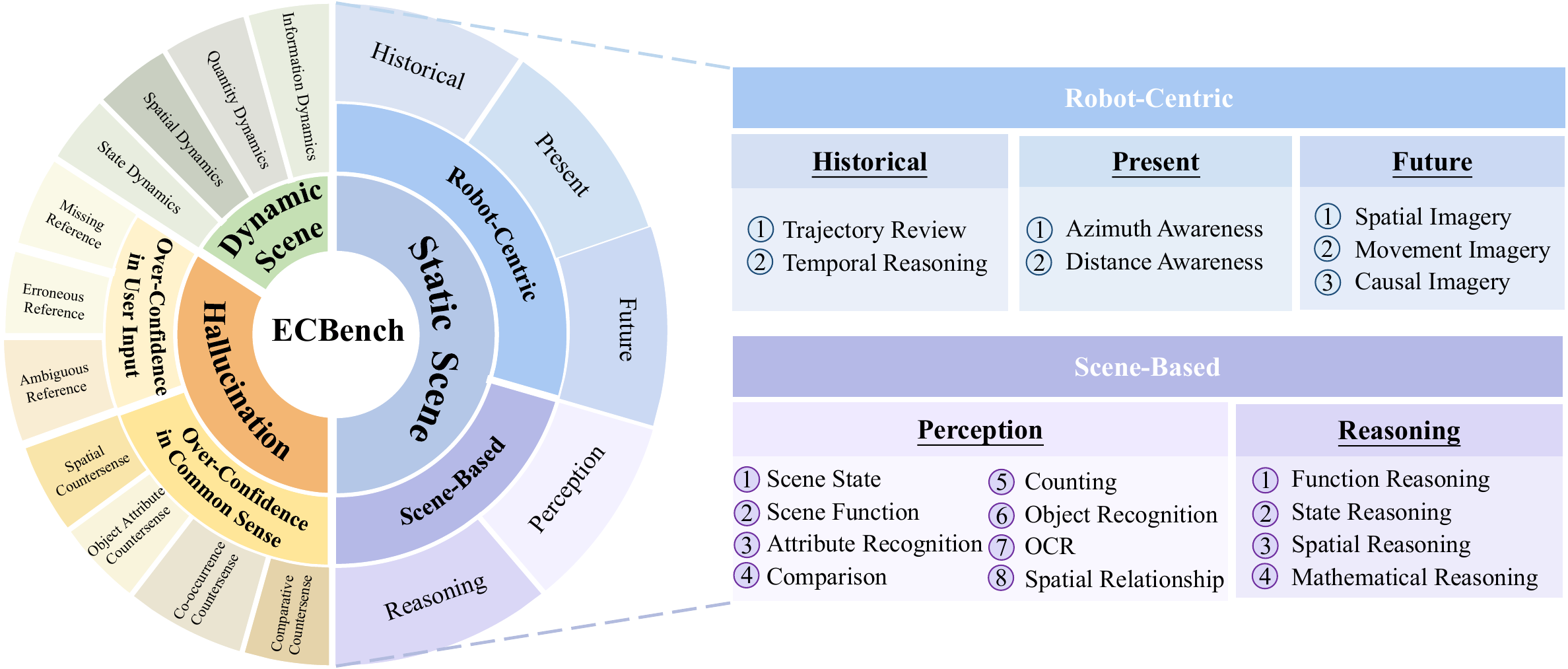}

   \caption{\textbf{Overview of embodied cognition dimensions in ECBench.} ECBench includes three subsets: static scenes, dynamic scenes, and hallucination, evaluating a total of 30 embodied cognitive abilities.}
   \vspace{-7pt}
   \label{fig:dimension}
\end{figure*}

In this section, we introduce the construction and detailed statistics of ECBench.

\subsection{Benchmark Construction}
\label{sec:Benchmark Construction}
\paragraph{Video Collection}

Traditional embodied question answering videos typically involve the comprehensive scanning of 3D scenes. Alternatively, some datasets, like OpenEQA and Env-QA, create more diverse videos by manually simulating motion trajectories in virtual scenes \cite{ai2thor,procthor}. In contrast, ECBench enhances video diversity by integrating virtual scenes with real scenes and combining static scenes with dynamic scenes. Although current LVLMs can't process depth information, ECBench still includes depth maps that are crucial for embodied tasks.

Firstly, we employ an open-world object navigation agent \cite{openfmnav} and an active embodied question-answering agent \cite{explore-eqa} to capture robotic authentic video streams in the HM3D environment \cite{hm3d}. These videos exhibit increased instances of hesitation and pauses compared to those manually gathered by humans, thereby offering a more realistic representation of robotic perception during task execution. Secondly, we carefully select 191 real scan videos from ScanNet \cite{scannet} and MultiScan \cite{multiscan}. Lastly, utilizing the Intel RealSense depth camera, we collect videos of counterintuitive scenes and dynamic scenes from the real world, providing a data basis for evaluating hallucination problems and dynamic perception.

\vspace{-0.8em}
\paragraph{Capability Taxonomy}
Motivated by the popular general VideoQA benchmarks \cite{mmbench-video,q-bench-video,egoschema,mvbench}, we implement an enhanced systematic approach to categorize various embodied cognitive abilities (Fig.~\ref{fig:dimension}). ECBench is categorized into three primary sets: static scene QA, dynamic scene QA, and hallucination evaluation. 

Static scene QA employs a hierarchical classification system divided into three levels of cognitive abilities. At the L1 level, there are two higher-order cognitive functions: scene-based cognition and robot-centric cognition. Scene-based cognition is concerned with understanding scene information and can be further divided into perception and reasoning. In contrast, robot-centric cognition necessitates a level of self-awareness in the model, enabling it to understand its position and impact within the physical environment. We divide robot-centric cognition into three parts: historical cognition, present cognition, and future cognition. Overall, static scene QA encompasses 19 distinct cognitive abilities, comprehensively covering the cognitive demands of robots in static environments. The detailed classification is presented in Fig. \ref{fig:dimension}.


\begin{figure*}[t]
  \centering
  \begin{subfigure}{0.3\linewidth}
   \includegraphics[width=0.99\linewidth]{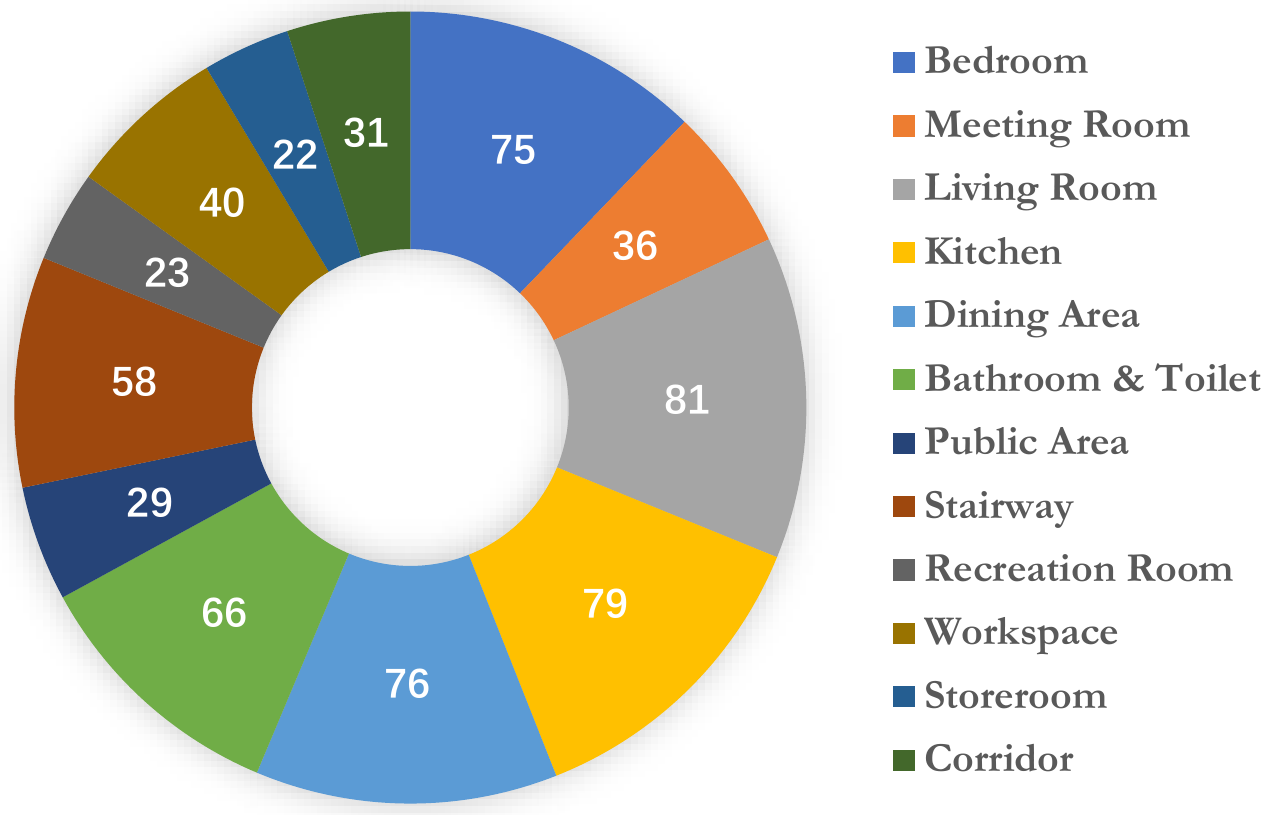}
   \caption{Number of various scenario categories.}
   \label{fig:scene_type}
  \end{subfigure}
  \hfill
  \begin{subfigure}{0.3\linewidth}
    \includegraphics[width=0.92\linewidth]{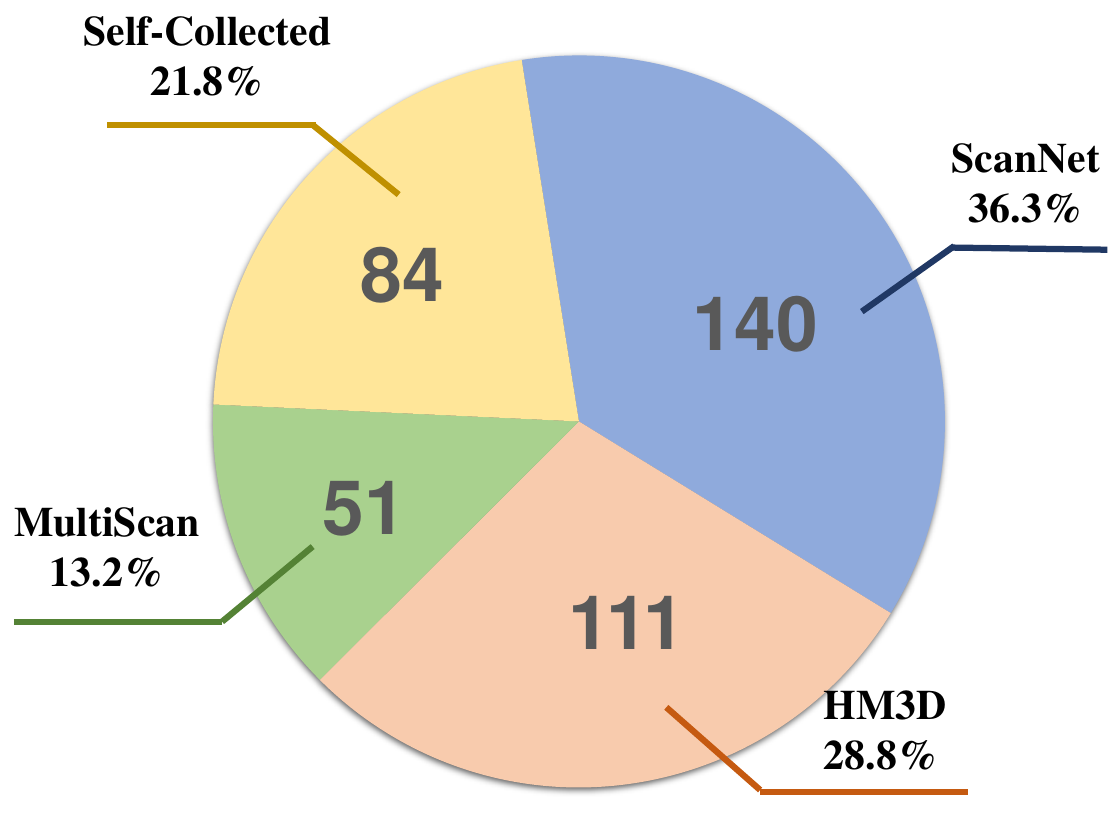}
   \caption{Distribution of video sources}
   \label{fig:video_source}
  \end{subfigure}
  \hfill
  \begin{subfigure}{0.35\linewidth}
  \includegraphics[width=0.95\linewidth]{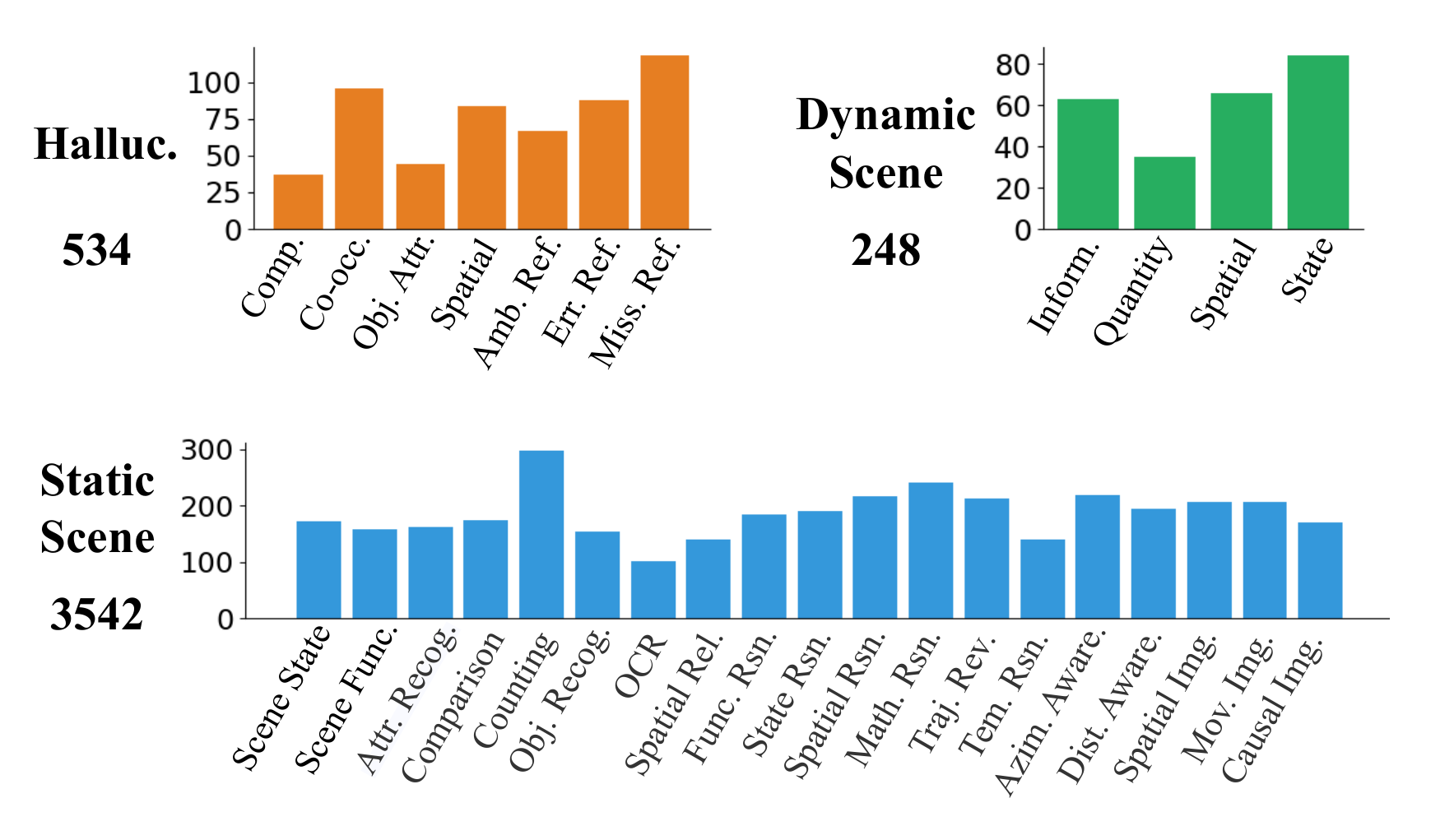}
   \caption{Distribution of QA pairs requiring different cognitive abilities.}
   \label{fig:qa_pair_num}
   \end{subfigure}
  \caption{\textbf{Data analysis of ECBench} reflects a rich diversity of scenario categories, video sources, and evaluation dimensions.}
  \vspace{-7pt}
  \label{fig:data analysis}
\end{figure*}

\begin{figure}[t]
  \centering
   \includegraphics[width=0.99\linewidth]{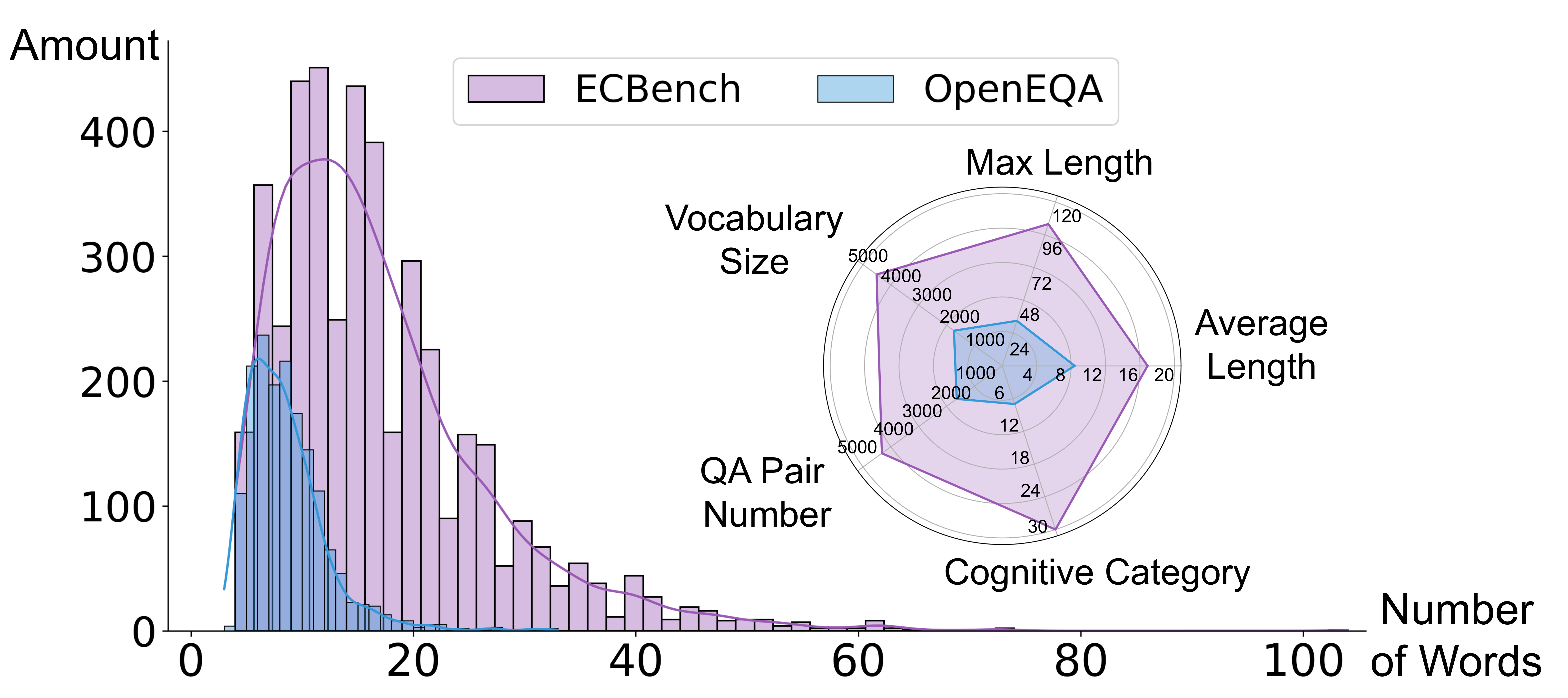}

   \caption{\textbf{Comparison with OpenEQA \cite{openeqa} on textual data}, including the distribution of question lengths, average question length, maximum question length, vocabulary size, number of questions, and number of capabilities.}
   \vspace{-10pt}
   \label{fig:qa_len}
\end{figure}

Dynamic scene QA can be divided into four categories according to their dynamic characteristics: spatial, state, information, and quantity dynamics. We design videos and questions specifically tailored to highlight these dynamic features. Emphasis is also placed on the temporal and spatial dynamics within robot movement in our questions.

In embodied scenarios, hallucinations are not merely a bias under 2D vision but are also closely related to the model's over-confidence in common sense and user input in 3D scenarios. We propose four categories of counterintuitive questions (Fig.~\ref{fig:dimension}) to evaluate LVLMs' commonsense hallucinations. Furthermore, over-confidence in user input can inhibit the model's ability to identify ambiguous references, missing references, and erroneous references in user instructions, significantly impairing the effectiveness of human-robot interaction.

Examples of each category are presented in Appendix~\ref{sec:Capability Taxonomy}.

\vspace{-0.8em}
\paragraph{Construction of Question-Answer Pairs}

Current benchmarks for embodied VideoQA utilize two forms of QA pair construction. One is to use template sentences and object annotations within 3D environments to automatically create standard QA pairs, which tend to have a limited textual form \cite{env-qa}. Another is to use LVLMs like GPT-4 to generate QA pairs, supplemented by automatic or manual filtering mechanisms \cite{robovqa}. However, these generated QA pairs only rely on single-frame information, lacking the ability to evaluate the model’s cognitive ability to entire 3D scenes and dynamic events. Therefore, we employ manual annotation methods to construct ECBench.

ECBench contains 30 cognitive abilities, which will result in a significant category imbalance if traditional video-by-video question-annotation methods are used. 
Therefore, we implement a category-independent annotation approach by assigning volunteers a predetermined number of tasks related to each cognitive ability. This method ensures a balanced distribution between the quantity of QA pairs associated with both rare and common abilities. 


After completing all data labeling, we conduct a filtering process for commonsense questions.
It is widely recognized that LVLMs possess extensive prior knowledge in common sense but often struggle to align with visual reality \cite{r-bench, throne, phd}. Our objective is to reduce the number of questions that rely solely on common sense,
allowing ECBench to concentrate on visual cognition. Specifically, we employ GPT-4o to answer all the questions six times without visual inputs, followed by manual revisions of questions consistently answered correctly to enhance their reliance on visual input. This process is iterated thrice. Ultimately, we perform stringent cross-validation on all questions to guarantee the accuracy of QA pairs.
The final ECBench includes static scene QA, dynamic scene QA, and hallucination QA, systematically evaluating LVLMs from various embodied cognitive abilities. 

\vspace{-0.8em}
\paragraph{Evaluation Framework}
In ECBench, most answers are close-ended and explicit, while some open-ended answers involve an element of ambiguity.
To address this issue, we propose ECEval, a combined evaluation system integrating both multi-level and binary judgments. 
We first partition ECBench into two parts: open-ended and close-ended. For the close-ended part, we prompt GPT-4o to assign a straightforward binary score of either 0 or 1. For the open-ended part, we manually annotate a \textit{0.5-point answer} for each open-ended question. In the evaluation, open-ended answers are scored by GPT-4o on a scale from 0 to 1 in increments of 0.2, based on \textit{0.5-point answers}. This evaluation framework facilitates more precise scoring for questions with definitive answers and provides a more nuanced scoring methodology for open-ended inquiries. Further details can be found in Appendix~\ref{sec:Evaluation Framework}.

\begin{table*}[t]
\centering
\small
\setlength\tabcolsep{5pt} 
\begin{tabular}{lcccccccccccc}
\Xhline{2\arrayrulewidth}
\multicolumn{1}{l|}{\multirow{2}{*}{\textbf{Model}}} & \multicolumn{1}{c|}{\multirow{2}{*}{\begin{tabular}[c]{@{}c@{}}\textbf{Overall}\\ \textbf{Mean}\end{tabular}}} & \multicolumn{3}{c|}{\textbf{Static Scene}} & \multicolumn{5}{c|}{\textbf{Dynamic Scene}} & \multicolumn{3}{c}{\textbf{Hallucination}} \\ \cline{3-13}
\multicolumn{1}{l|}{} & \multicolumn{1}{c|}{} & \textbf{SB} & \multicolumn{1}{c|}{\textbf{RC}} & \multicolumn{1}{c|}{\textbf{Mean}} & \textbf{ID} & \textbf{QD} & \textbf{SPD} & \multicolumn{1}{c|}{\textbf{STD}} & \multicolumn{1}{c|}{\textbf{Mean}} & \textbf{UI} & \multicolumn{1}{c|}{\textbf{CS}} & \textbf{Mean} \\ \Xhline{2\arrayrulewidth}
\rowcolor{mygray!20}
\multicolumn{13}{c}{\textit{Blind LLMs}} \\ \Xhline{2\arrayrulewidth}
\multicolumn{1}{l|}{GPT-4o} & \multicolumn{1}{c|}{24.09} & \multicolumn{1}{c}{25.32} & \multicolumn{1}{c|}{33.06} & \multicolumn{1}{c|}{28.26} & \multicolumn{1}{c}{4.76} & \multicolumn{1}{c}{0.00} & \multicolumn{1}{c}{10.00} & \multicolumn{1}{c|}{13.81} & \multicolumn{1}{c|}{8.55} & \multicolumn{1}{c}{1.83} & \multicolumn{1}{c|}{5.36} & \multicolumn{1}{c}{3.56} \\ \Xhline{2\arrayrulewidth}
\rowcolor{myred!10}
\multicolumn{13}{c}{\textit{Native Video-LVLMs}} \\ \Xhline{2\arrayrulewidth}
\multicolumn{1}{l|}{Video-LLaMA2-7B-[16f] \cite{VideoLLaMA2}} & \multicolumn{1}{c|}{33.87} & 41.23 & \multicolumn{1}{c|}{33.52} & \multicolumn{1}{c|}{38.30} &16.51  & 5.71 & 16.97 & \multicolumn{1}{c|}{11.19} & \multicolumn{1}{c|}{13.31} & 4.03 & \multicolumn{1}{c|}{24.29} & 13.93 \\
\multicolumn{1}{l|}{Video-LLaVA-7B-[8f] \cite{videollava}} & \multicolumn{1}{c|}{35.25} & 41.21 & \multicolumn{1}{c|}{37.25} & \multicolumn{1}{c|}{39.71} & 18.41 & 14.29 & \textbf{19.20} & \multicolumn{1}{c|}{\textbf{16.67}} & \multicolumn{1}{c|}{17.66} & 1.10 & \multicolumn{1}{c|}{26.97} & 13.75 \\
\multicolumn{1}{l|}{Kangaroo-8B-[64f] \cite{kangaroo}} & \multicolumn{1}{c|}{36.23} & 40.79  & \multicolumn{1}{c|}{\textbf{40.01}} & \multicolumn{1}{c|}{40.49} & 19.20 & \textbf{17.14} & 16.97 & \multicolumn{1}{c|}{15.95} & \multicolumn{1}{c|}{17.42} & 0.37 & \multicolumn{1}{c|}{\textbf{33.49}} & 16.55 \\ 
\multicolumn{1}{l|}{LongVA-7B-[384f] \cite{longva}} & \multicolumn{1}{c|}{\textbf{40.47}} & \textbf{49.03} & \multicolumn{1}{c|}{38.56} & \multicolumn{1}{c|}{\textbf{45.05}} & \textbf{28.57} & 8.57 & 13.94 & \multicolumn{1}{c|}{\textbf{16.67}} & \multicolumn{1}{c|}{\textbf{17.82}} & \textbf{8.06} & \multicolumn{1}{c|}{\textbf{33.49}} & \textbf{20.49} \\
\Xhline{2\arrayrulewidth}
\rowcolor{myyellow!30}
\multicolumn{13}{c}{\textit{Open-Source Image-LVLMs}} \\ \Xhline{2\arrayrulewidth}
\multicolumn{1}{l|}{Idefics3-8B-[8f] \cite{laurenccon2024building}}  & \multicolumn{1}{c|}{32.76} & 36.31 & \multicolumn{1}{c|}{36.49} & \multicolumn{1}{c|}{36.38} & 11.43 & 8.57 & 13.64 & \multicolumn{1}{c|}{14.53} & \multicolumn{1}{c|}{12.66} & \textbf{5.86} & \multicolumn{1}{c|}{30.73} & 18.01 \\
\multicolumn{1}{l|}{Idefics3-8B-[20f]} & \multicolumn{1}{c|}{33.35} & 36.56 & \multicolumn{1}{c|}{38.09} & \multicolumn{1}{c|}{37.14} & 18.10 & 8.57 & 10.61 & \multicolumn{1}{c|}{19.29} & \multicolumn{1}{c|}{15.16} & 4.76 & \multicolumn{1}{c|}{28.89} & 16.55 \\

\multicolumn{1}{l|}{InternVL2-40B-[8f] \cite{internvl2}} & \multicolumn{1}{c|}{35.41} & 40.34 & \multicolumn{1}{c|}{38.16} & \multicolumn{1}{c|}{39.51} & 30.16 & 2.86 & 15.15 & \multicolumn{1}{c|}{16.90} & \multicolumn{1}{c|}{17.82} & 2.93 & \multicolumn{1}{c|}{30.19} & 16.25 \\
\multicolumn{1}{l|}{InternVL2-40B-[20f]} & \multicolumn{1}{c|}{35.94} & 41.27 & \multicolumn{1}{c|}{38.09} & \multicolumn{1}{c|}{40.06} & 33.33 & 0.00 & 12.12 & \multicolumn{1}{c|}{21.66} & \multicolumn{1}{c|}{19.03} & 3.66 & \multicolumn{1}{c|}{29.58} & 16.33 \\
\multicolumn{1}{l|}{Qwen2VL-7B-[8f] \cite{qwen2Vl}}  & \multicolumn{1}{c|}{37.28} & 43.23 & \multicolumn{1}{c|}{38.31} & \multicolumn{1}{c|}{41.36} & 23.81 & 5.71 & 19.67 & \multicolumn{1}{c|}{16.90} & \multicolumn{1}{c|}{17.82} & 5.49 & \multicolumn{1}{c|}{33.33} & 19.10 \\ 
\multicolumn{1}{l|}{Qwen2VL-7B-[20f]} & \multicolumn{1}{c|}{39.57} & 46.21 & \multicolumn{1}{c|}{39.33} & \multicolumn{1}{c|}{43.60} & 32.70 & 8.57 & \textbf{19.70} & \multicolumn{1}{c|}{18.33} & \multicolumn{1}{c|}{20.97} & 4.40 & \multicolumn{1}{c|}{39.08} & 21.35 \\ 
\multicolumn{1}{l|}{Qwen2VL-72B-[8f]} & \multicolumn{1}{c|}{41.57} & 48.28 & \multicolumn{1}{c|}{43.52} & \multicolumn{1}{c|}{46.47} & 33.65 & 8.57 & 18.18 & \multicolumn{1}{c|}{16.43} & \multicolumn{1}{c|}{20.16} & 4.76 & \multicolumn{1}{c|}{33.56} & 18.84 \\ 
\multicolumn{1}{l|}{Qwen2VL-72B-[20f]} & \multicolumn{1}{c|}{\textbf{44.62}} & \textbf{52.40} & \multicolumn{1}{c|}{\textbf{43.95}} & \multicolumn{1}{c|}{\textbf{49.19}} & \textbf{37.14} & \textbf{11.43} & 18.18 & \multicolumn{1}{c|}{\textbf{24.05}} & \multicolumn{1}{c|}{\textbf{24.03}} & \textbf{5.86} & \multicolumn{1}{c|}{\textbf{42.38}} & \textbf{23.71} \\ \Xhline{2\arrayrulewidth}
\rowcolor{mygreen!50}
\multicolumn{13}{c}{\textit{Proprietary Image-LVLMs}} \\ \Xhline{2\arrayrulewidth}
\multicolumn{1}{l|}{GPT-4v-[8f]  \cite{gpt4}} & \multicolumn{1}{c|}{39.16} & 45.09 & \multicolumn{1}{c|}{40.16} & \multicolumn{1}{c|}{43.22} & 29.52 & 8.57 & 22.12 & \multicolumn{1}{c|}{20.24} & \multicolumn{1}{c|}{21.45} & 9.16 & \multicolumn{1}{c|}{32.11} & 20.37 \\
\multicolumn{1}{l|}{GPT-4o-mini-[8f] \cite{gpt4o}} & \multicolumn{1}{c|}{39.71} & 46.00 & \multicolumn{1}{c|}{40.35} & \multicolumn{1}{c|}{43.86} & 28.25 & 17.14 & 16.97 & \multicolumn{1}{c|}{19.05} & \multicolumn{1}{c|}{20.56} & 9.16 & \multicolumn{1}{c|}{33.26} & 20.94 \\
\multicolumn{1}{l|}{GPT-4o-mini-[32f]} & \multicolumn{1}{c|}{43.69} & 51.20 & \multicolumn{1}{c|}{43.12} & \multicolumn{1}{c|}{48.13} & 26.35 & \textbf{20.00} & 15.15 & \multicolumn{1}{c|}{18.10} & \multicolumn{1}{c|}{19.68} & \textbf{9.89} & \multicolumn{1}{c|}{41.38} & 25.28 \\
\multicolumn{1}{l|}{GPT-4o-[8f] \cite{gpt4o}} & \multicolumn{1}{c|}{45.96} & 52.95 & \multicolumn{1}{c|}{45.91} & \multicolumn{1}{c|}{50.27} & \textbf{36.83} & 11.43 & 21.21 & \multicolumn{1}{c|}{23.33} & \multicolumn{1}{c|}{\textbf{24.52}} & 9.52 & \multicolumn{1}{c|}{45.67} & 27.19 \\ 
\multicolumn{1}{l|}{GPT-4o-[32f]} & \multicolumn{1}{c|}{\textbf{50.35}} & \textbf{59.74} & \multicolumn{1}{c|}{\textbf{49.04}} & \multicolumn{1}{c|}{\textbf{55.06}} & 25.71 & 14.29 & \textbf{22.73} & \multicolumn{1}{c|}{\textbf{24.04}} & \multicolumn{1}{c|}{22.74} & 7.69 & \multicolumn{1}{c|}{\textbf{57.01}} & \textbf{31.80} \\ \Xhline{2\arrayrulewidth}
\rowcolor{myblue}
\multicolumn{13}{c}{\textit{Embodied / Egocentric LVLMs}} \\ \Xhline{2\arrayrulewidth}
\multicolumn{1}{l|}{GeLM-7B-[180f] \cite{multihop-egoqa}} & \multicolumn{1}{c|}{21.54} & 25.72 & \multicolumn{1}{c|}{23.70} & \multicolumn{1}{c|}{24.95} & 5.08 & 5.71 & 8.18 & \multicolumn{1}{c|}{3.51} & \multicolumn{1}{c|}{5.48} & 0.37 & \multicolumn{1}{c|}{12.49} & 6.29 \\
\multicolumn{1}{l|}{AlanaVLM-7B-[64f] \cite{evud}} & \multicolumn{1}{c|}{\textbf{34.75}} & \textbf{40.38} & \multicolumn{1}{c|}{\textbf{36.61}} & \multicolumn{1}{c|}{\textbf{38.95}} & \textbf{19.05} & \textbf{5.71} & \textbf{19.70} & \multicolumn{1}{c|}{\textbf{14.76}} & \multicolumn{1}{c|}{\textbf{15.89}} & \textbf{1.83} & \multicolumn{1}{c|}{\textbf{29.96}} &  \textbf{15.58}\\ \Xhline{2\arrayrulewidth}
\rowcolor{blue!8}
\multicolumn{1}{l|}{Human} & \multicolumn{1}{c|}{94.96} & \multicolumn{1}{c}{97.45} & \multicolumn{1}{c|}{93.21} & \multicolumn{1}{c|}{95.86} & \multicolumn{1}{c}{91.77} & \multicolumn{1}{c}{97.21} & \multicolumn{1}{c}{94.32} & \multicolumn{1}{c|}{96.25} & \multicolumn{1}{c|}{94.55} & \multicolumn{1}{c}{82.74} & \multicolumn{1}{c|}{96.91} & \multicolumn{1}{c}{89.21} \\ \Xhline{2\arrayrulewidth}
\end{tabular}
\caption{\textbf{Main evaluation results of various LVLMs on ECBench.} We evaluate in three major categories: Static Scenes, Dynamic Scenes, and Hallucinations. SB and RC represent Scene-Based and Robot-Centric, respectively. ID, QD, SPD, and STD stand for Information Dynamics, Quantity Dynamics, Spatial Dynamics, and State Dynamics, respectively. UI and CS represent Over-Confidence in User Input and Over-Confidence in Common Sense, respectively. -[Nf] indicates the method take N frames uniformly sampled from a video as input.}
\vspace{-7pt}
\label{tab:main_result}
\end{table*}

\subsection{Dataset Statistics}
\label{sec: Dataset Statistics}
\paragraph{Video Source}
As illustrated in Fig.~\ref{fig:video_source}, ECBench comprises 386 RGBD videos exploring indoor scenes, sourced from ScanNet \cite{scannet}, MultiScan \cite{multiscan}, HM3D \cite{hm3d}, and self-collections. We select 140 and 51 real-world videos from the ScanNet and MultiScan datasets, respectively, chosen based on the richness of scene information. Additionally, ECBench includes 111 first-person perspective videos of agents performing tasks in the HM3D virtual environment, with 49 videos originating from the Embodied Question Answering Agent \cite{explore-eqa} and 62 videos from the Object Navigation Agent \cite{openfmnav}. 
We manually collect scenes that are less common in public datasets, featuring 44 counterintuitive scenes and 40 dynamic scenes. As depicted in Fig.~\ref{fig:scene_type}, each video's scene category has been manually annotated. 
Overall, ECBench comprises 12 distinct indoor scenes. Beyond common scenes like bedrooms and kitchens, ECBench also includes rare scenes such as stairways and recreation rooms, providing a comprehensive evaluation of the model’s adaptability across diverse environments.

\vspace{-0.8em}
\paragraph{QA Pairs}
As shown in Fig.~\ref{fig:qa_pair_num}, ECBench comprises 4,324 QA pairs, systematically evaluating LVLMs from 30 cognitive perspectives. In previous benchmarks, such as OpenEQA \cite{openeqa}, the QA pairs only include a limited range of ability categories. ECBench not only provides a more systematic classification and analysis of embodied cognitive abilities but also introduces 534 hallucination QA pairs and 248 dynamic scene QA pairs. As illustrated in Fig.~\ref{fig:qa_len}, the average question length in ECBench is 16.88, which is twice that of OpenEQA. ECBench includes more complex and diverse questions and contains 
a vocabulary of 4,513 words, which is 2.6 times larger than OpenEQA. Further dataset analysis is presented in Appendix~\ref{sec:Additional Dataset Analysis}.

%% file: sec/4_Experiment.tex
\section{Experiment}
Based on ECBench, we conduct extensive evaluations of native Video-LVLMs, open-source and proprietary Image-LVLMs, as well as embodied and egocentric LVLMs. For the open-source models, we utilize their default hyperparameters for inference. Further experimental details and prompts are presented in Appendix~\ref{sec:More Details of Experiment}.

\subsection{Main Results}
\paragraph{Blind LLM}
In the first row of Tab.~\ref{tab:main_result}, we conduct a blind evaluation of ECBench using GPT-4o \cite{gpt4o} without the video input. This score reflects the extent to which reliance solely on the common sense and guessing ability of LVLMs can achieve. ECBench includes 90.31\% close-ended questions to ensure the accuracy of the evaluation. Despite this, the blind evaluation score for ECBench reaches only 24.09, which is a 28\% reduction compared to the entirely open-ended OpenEQA (33.5). This indicates that ECBench significantly outperforms OpenEQA in terms of evaluation precision and visual dependency.

\vspace{-0.8em}
\paragraph{Native Video-LVLMs}
We assess popular general and long context native Video-LVLMs. The long video models are capable of incorporating more frames, which is critical for embodied videos with high dynamic views. Consequently, the SOTA model in the domain of long video understanding, LongVA-[384f] \cite{longva}, achieves 40.47, making it the best among open-source Video-LVLMs. However, in dynamic scenes, models with more input frames do not show a significant improvement. This suggests that, Video-LVLMs fundamentally lacks the ability to capture and understand dynamic scenes.

\begin{table*}[t]
\centering
\small
\setlength\tabcolsep{3pt} 
\begin{tabular}{l|c|ccccccccc|ccccc}
\Xhline{2\arrayrulewidth}
\multirow{2}{*}{\textbf{Model}} & \multirow{2}{*}{\begin{tabular}[c]{@{}c@{}}\textbf{Overall}\\ 
\textbf{Mean}\end{tabular}} & \multicolumn{9}{c|}{\textbf{Perception}} & \multicolumn{5}{c}{\textbf{Reasoning}} \\ \cline{3-16}
 &  & \textbf{SS} & \textbf{SF} & \textbf{AR} & \textbf{COM} & \textbf{COU} & \textbf{OR} & \textbf{OCR} & \multicolumn{1}{c|}{\textbf{SR}} & \textbf{Mean} & \textbf{FR} & \textbf{STR} & \textbf{SPR} & \multicolumn{1}{c|}{\textbf{MR}} & \textbf{Mean} \\ \Xhline{2\arrayrulewidth}
AlanaVLM-7B-{[}64f{]} & 40.38 & 52.60 & 59.37 & 46.21 & 39.08 & 28.75 & 52.21 & 7.84 & \multicolumn{1}{c|}{34.68} & 40.44 & 55.46 & 49.46 & 39.54 & \multicolumn{1}{c|}{21.49} & 40.28 \\
LongVA-7B-{[}384f{]} & 49.03 & \textbf{71.10} & 74.05 & 62.61 & 51.15 & 32.19 & 60.39 & 31.76 & \multicolumn{1}{c|}{43.17} & 52.34 & \textbf{61.84} & 56.39 & 40.00 & \multicolumn{1}{c|}{22.41} & 43.73 \\
Qwen2VL-72B-{[}8f{]} & 48.28 & 64.74 & 75.57 & 67.45 & 36.78 & 38.59 & 55.97 & 19.61 & \multicolumn{1}{c|}{42.30} & 
50.34 & 43.14 & 55.41 & 42.27 & \multicolumn{1}{c|}{36.35} & 44.98 \\
Qwen2VL-72B-{[}20f{]} & 52.40 & 64.05 & 74.30 & 71.68 & 35.06 & 48.69 & 65.58 & 43.73 & \multicolumn{1}{c|}{37.99} & 55.05 & 51.68 & \textbf{60.20} & 48.66  & \multicolumn{1}{c|}{34.77} & 48.16 \\
GPT-4o-{[}8f{]} & 52.95 & 65.20 & 74.68 & 72.05 & 59.77 & 43.30 & 65.58 & 24.51 & \multicolumn{1}{c|}{47.48} & 56.80 & 48.43 & 51.02 & 47.00 & \multicolumn{1}{c|}{41.66} & 46.77 \\
GPT-4o-{[}32f{]} & \textbf{59.74} & 67.98 & \textbf{77.97}& \textbf{75.03} & \textbf{61.49} & \textbf{54.21} & \textbf{72.21} & \textbf{49.02} & \multicolumn{1}{c|}{\textbf{47.91}} & \textbf{63.14} & 50.81 & 54.21 & \textbf{49.49} & \multicolumn{1}{c|}{\textbf{49.54}} & \textbf{51.70} \\ \Xhline{2\arrayrulewidth}
\end{tabular}
\caption{\textbf{Evaluation of sub-abilities under the Scene-Based (SB) category in the static scene set.} SS: Scene State, SF: Scene Function, AR: Attribute Recognition, COM: Comparison, COU: Counting, OR: Object Recognition, SR: Spatial Relationship. FR: Function Reasoning, STR: State Reasoning, SPR: Spatial Reasoning, MR: Mathematical Reasoning.}
\label{tab:scene-based}
\end{table*}

\begin{table*}[t]
\centering
\small
\setlength\tabcolsep{7.4pt} 
\begin{tabular}{l|c|ccc|ccc|cccc}
\Xhline{2\arrayrulewidth}
\multirow{2}{*}{\textbf{Model}} & \multirow{2}{*}{\begin{tabular}[c]{@{}c@{}}\textbf{Overall}\\ \textbf{Mean}\end{tabular}} & \multicolumn{3}{c|}{\textbf{Historical}} & \multicolumn{3}{c|}{\textbf{Present}} & \multicolumn{4}{c}{\textbf{Future}} \\ \cline{3-12} 
 &  & \textbf{TJR} & \multicolumn{1}{c|}{\textbf{TPR}} & \textbf{Mean} & \textbf{AA} & \multicolumn{1}{c|}{\textbf{DA}} & \textbf{Mean} & \textbf{SI} & \textbf{MI} & \multicolumn{1}{c|}{\textbf{CI}} & \textbf{Mean} \\ \Xhline{2\arrayrulewidth}
AlanaVLM-7B-{[}64f{]} & 36.61 & 40.09 & \multicolumn{1}{c|}{37.14} & 38.92 & 33.61 & \multicolumn{1}{c|}{34.36} & 33.96 & 38.65 & 33.72 & \multicolumn{1}{c|}{39.30} & 37.09 \\
LongVA-7B-{[}384f{]} & 38.56 & 40.47 & \multicolumn{1}{c|}{36.43} & 38.86 & 37.81 & \multicolumn{1}{c|}{41.03} & 39.32 & 35.27 & 36.52 & \multicolumn{1}{c|}{42.57} & 37.85 \\
Qwen2VL-72B-{[}8f{]} & 43.52 & 43.96 & \multicolumn{1}{c|}{50.00} & 46.36 & 34.70 & \multicolumn{1}{c|}{43.08} & 38.65 & 43.67 & 43.67 & \multicolumn{1}{c|}{49.12} & 45.26 \\
Qwen2VL-72B-{[}20f{]} & 43.95 & 44.34 & \multicolumn{1}{c|}{48.57} & 46.02 & \textbf{39.00} & \multicolumn{1}{c|}{41.54} & 40.19 & 43.86 & 45.12 & \multicolumn{1}{c|}{47.49} & 45.37 \\
GPT-4o-{[}8f{]} & 45.91 & 47.26 & \multicolumn{1}{c|}{\textbf{54.29}} & 50.06 & 37.44 & \multicolumn{1}{c|}{46.67} & 41.79 & 42.03 & 46.09 & \multicolumn{1}{c|}{51.81} & 46.32 \\
GPT-4o-{[}32f{]} & \textbf{49.04} & \textbf{52.17} & \multicolumn{1}{c|}{53.57} & \textbf{52.73} & 38.17 & \multicolumn{1}{c|}{\textbf{51.28}} & \textbf{44.35} & \textbf{47.92} & \textbf{47.63} & \multicolumn{1}{c|}{\textbf{55.91}} & \textbf{50.15} \\ \Xhline{2\arrayrulewidth}
\end{tabular}
\caption{\textbf{Evaluation of sub-abilities under the Robot-Centric (RC) category in the static scene set.} TJR: Trajectory Review, TPR: Temporal Reasoning, AA: Azimuth Awareness, DA: Distance Awareness, SI: Spatial Imagery, MI: Movement Imagery, CI: Causal Imagery. }
\vspace{-10pt}
\label{tab:robot-centric}
\end{table*}

\vspace{-0.8em}
\paragraph{Open-Source Image-LVLMs}
Qwen2VL \cite{qwen2Vl} exhibited impressive performance, achieving 44.16, surpassing GPT-4o-mini. Additionally, we observe that as model capabilities improved, the gains from increasing the number of input frames became more pronounced. When the input is expanded from 8 frames to 20 frames, the Qwen2VL-72B score increases by 3.05, whereas the performance of InternVL2-40B \cite{internvl2} remains virtually unchanged.

\vspace{-0.8em}
\paragraph{Proprietary Image-LVLMs}
The GPT series models are widely regarded as the most capable proprietary models for multimodal understanding. We evaluate several models from this series, including GPT-4v \cite{gpt4}, GPT-4o-mini, and GPT-4o. GPT-4o-[32f] exhibits the strongest embodied cognitive abilities, achieving 50.35. This score is 12.8\% higher than the best open-source Image-LVLMs and 24.4\% higher than the best native Video-LVLMs.

\vspace{-0.8em}
\paragraph{Embodied / Egocentric LVLMs}
Currently, there are several LVLMs designed to handle multimodal understanding in embodied and egocentric contexts. We evaluate two notable models: AlanaVLM \cite{evud} and GeLM \cite{multihop-egoqa}. Unfortunately, their performance in embodied cognition is underwhelming. GeLM, which is mainly trained using egocentric videos obtained from XR devices, faced a significant domain gap when applied to robotic videos, resulting in a low score of 21.54. AlanaVLM performs reasonably well on the OpenEQA benchmark, but it scores only 34.75 on ECBench. These findings suggest that ECBench offers more thorough evaluation dimensions, objectively reflecting the embodied cognitive capabilities of LVLMs.

\subsection{Evaluation in Static Scene}
Questions in static scenes are divided into two categories: Scene-Based and Robot-Centric. Scene-Based questions concentrate on assessing the LVLMs' scene comprehension from a third-person viewpoint, whereas Robot-Centric questions focus on the relationship between the robot itself and the environment from a first-person viewpoint.


\vspace{-0.8em}
\paragraph{Scene-Based Questions}

Table~\ref{tab:scene-based} illustrates the ability of various models to answer Scene-Based questions. Among the eight categories of Perception questions, Counting, OCR, and Spatial Relationship are the areas where current LVLMs are least proficient. These deficiencies hinder robots from accurately recognizing the intricacies of their surrounding environments when completing tasks. For GPT-4o-[32f], the scores for Attribute Recognition and Object Recognition are 75.03 and 72.21, respectively. This suggests that the advanced LVLMs can provide crucial cognition abilities for robots to navigate and manipulate individual objects. Among the four categories of Reasoning questions, Spatial Reasoning and Mathematical Reasoning provide greater challenges for the models. GPT-4o attains scores of 49.49 and 49.54 on these two difficult reasoning problems, clearly surpassing other models. LongVA and Qwen2VL excel in Function Reasoning and State Reasoning, achieving scores of 61.84 and 60.20, respectively. Overall, Reasoning problems impose higher demands on model capabilities compared to Perception questions. 

\vspace{-0.8em}
\paragraph{Robot-Centric Questions}

Table~\ref{tab:robot-centric} demonstrates the ability of various models to answer Robot-Centric questions, which are categorized as Historical, Present, and Future. As an embodied LVLM, AlanaVLM achieves a score of only 36.61 in the Robot-Centric cognition, significantly lagging behind the SOTA generalist LVLMs. Current embodied LVLMs do not acknowledge the significance of self-awareness in robotics. LongVA-7B-[384f], as a model designed for long video understanding, performs exceptionally well in the Present category, with only a 0.87 gap from Qwen2VL-72B. The richer temporal visual information is crucial for robots to construct scene models and develop self-awareness (e.g. position and orientation). Notably, the number of theoretically answerable questions for robot-centric questions significantly increases with the enhancement of the input frame count. However, it is only in powerful proprietary models like GPT-4o that we can clearly observe the benefits of increasing input frame numbers. 
Thus, current LVLMs urgently need to improve their ability to capture egocentric temporal information.

\begin{table}[t]
\small
\centering
\setlength\tabcolsep{3.5pt} 
\begin{tabular}{l|ccc|cccc}
\Xhline{2\arrayrulewidth}
\multirow{2}{*}{\textbf{Model}} & \multicolumn{3}{c|}{\textbf{UI}} & \multicolumn{4}{c}{\textbf{CS}} \\ \cline{2-8} 
 & \textbf{MRF} & \textbf{ERF} & \textbf{ARF} & \textbf{SC} & \textbf{OAC} & \textbf{COC} & \textbf{CC} \\ \Xhline{2\arrayrulewidth}
AlanaVLM & 1.69 & 3.41 & 0.00 & 28.33 & 24.09 & 28.96 & 43.24 \\
LongVA & 11.02 & \textbf{5.68} & \textbf{5.97} & 25.95 & 41.36 & 32.71 & 43.24 \\
Qwen2VL & 11.86 & 2.27 & 0.00 & \textbf{47.14} & 38.18 & 40.83 & 40.54 \\
GPT-4o & \textbf{14.41} & 4.55 & 0.00 & 45.95 & \textbf{55.45} & \textbf{66.46} & \textbf{59.46} \\ \Xhline{2\arrayrulewidth}
\end{tabular}
\caption{\textbf{Evaluation of sub-abilities in the Hallucination set.} MRF, ERF, and ARF represent Missing, Erroneous, and Ambiguous Reference. SC, OAC, COC, and CC represent Spatial, Object Attribute, Co-occurrence, and Comparative Countersense.}
\vspace{-10pt}
\label{tab:hallucination}
\end{table}

\subsection{Evaluation in Dynamic Scene}

All Dynamic Scene questions involve changes occurring within the scene. Notably, numerous changes occur outside the robot's immediate line of sight. LVLMs need to remember and compare scene information to effectively answer these questions. These questions are categorized into four types: Information Dynamics, Quantity Dynamics, Spatial Dynamics, and State Dynamics. As shown in Table~\ref{tab:main_result}, Quantity Dynamics presents the greatest challenge for LVLMs, with InternVL2-40B-[20f] even scoring zero on this task. Overall, current models show a complete inability to perceive the dynamic elements in the scene. However, for robots, the environment is constantly changing while they execute their tasks, which raises concerns about how to enable LVLMs to fully cognize these dynamics.  

\subsection{Evaluation of Hallucinatory Phenomena}
The primary hallucination issues when completing embodied tasks are categorized into two types: Over-Confidence in User Input and Over-Confidence in Common Sense. We hope that LVLMs can recognize erroneous or ambiguous user inputs and correct them through subsequent dialogue. However, in the UI (User Input) category of Table~\ref{tab:hallucination}, none of the LVLMs manages to handle this situation. In contrast, some questions in the CS (Common Sense) category are addressed with relatively good accuracy. Notably, GPT-4o achieves a score of 66.46 on co-occurrence countersense questions. This commendable performance is largely attributed to ongoing efforts in general multimodal hallucination research \cite{hallusionbench, pope, mitigating} regarding these problems. We hope that ECBench will draw the attention of more researchers to embodied hallucination issues and promote the development of more robust embodied cognitive LVLMs.

\begin{figure}[t]
  \centering
   \includegraphics[width=0.99\linewidth]{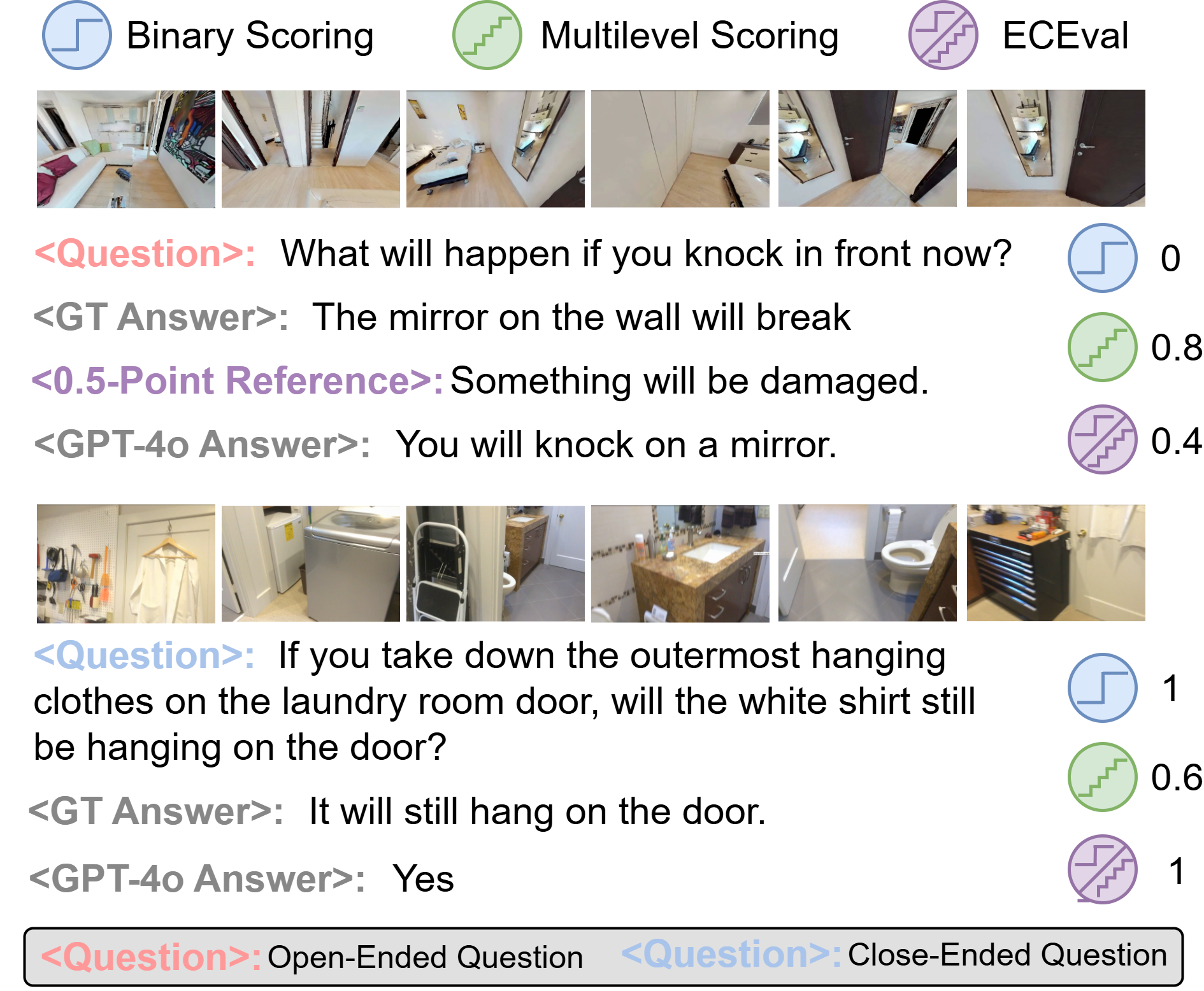}

   \caption{\textbf{Comparison of results between ECEval, Binary Scoring, and Multilevel Scoring}, for open-ended and closed-ended questions. Notably, only open-ended questions are annotated with \textit{0.5-point answers} .}
   \vspace{-7pt}
   \label{fig:ECEval}
\end{figure}

\subsection{ECEval Evaluation Method}
Currently, there are two mainstream evaluation methods for question-answering tasks in open-world settings. One is binary scoring, the other is multi-level scoring. In Fig.~\ref{fig:ECEval}, we emphasize the advantages of our ECEval method  relative to these two typical evaluation methods. As shown in the open-ended example in Fig.~\ref{fig:ECEval}, binary scoring fails to sensitively indicate changes in model abilities. Alternatively, the multi-level scoring without reference may result in inaccurate ratings due to GPT4o biases. ECEval has \textit{0.5-point answers} as references for open-ended questions, allowing GPT4o to provide more accurate scoring. As shown in the close-ended example in Fig.~\ref{fig:ECEval}, while the GPT4o answer differs in phrasing from the standard answer, the semantic content is entirely consistent. However, using multi-level scoring would prevent awarding full points. ECEval employs a binary evaluation mode for close-ended questions to avoid such issues. Thus, in a comprehensive benchmark like ECBench, relying solely on binary or multi-level evaluation modes is inherently flawed. Our proposed ECEval synthesizes the strengths of these evaluation methods, yielding more adaptable and precise evaluation outcomes. 

%% file: sec/5_Conclusion.tex
\section{Conclusion}
This work proposes ECBench, a holistic open-world benchmark for embodied cognition. ECBench comprises three test sets: static scene, dynamic scene, and hallucination, enabling a detailed assessment of LVLMs' capabilities across 30 embodied cognition tasks.
In addition, we introduce ECEval, which ensures the accuracy and smoothness of the ECBench scoring.
Extensive evaluations on ECBench reveal that various LVLMs struggle to perform well on common embodied situations, such as robot-centric, dynamic scenes and instruction ambiguity. We hope that ECBench will facilitate the advancement of LVLMs towards a more complex and diverse physical world.

%% file: sec/X_suppl.tex
\clearpage
\setcounter{page}{1}
\maketitlesupplementary


\section{Capability Taxonomy Details} 
\label{sec:Capability Taxonomy}

In Sec.~\ref{sec:Benchmark Construction} of the main paper and in Fig.~\ref{fig:dimension}, we present the process of constructing the embodied cognition dimensions for ECBench. In this section, we will elaborate on the design principles and specific details underlying each capability dimension evaluated within ECBench. Furthermore, we will provide numerous examples for reference.

\subsection{Static Scene}
In static scene datasets, there are two types of questions: scene-based and robot-centric.

\subsubsection{Scene-Based}

\begin{figure*}[t]
  \centering
   \includegraphics[width=0.9\linewidth]{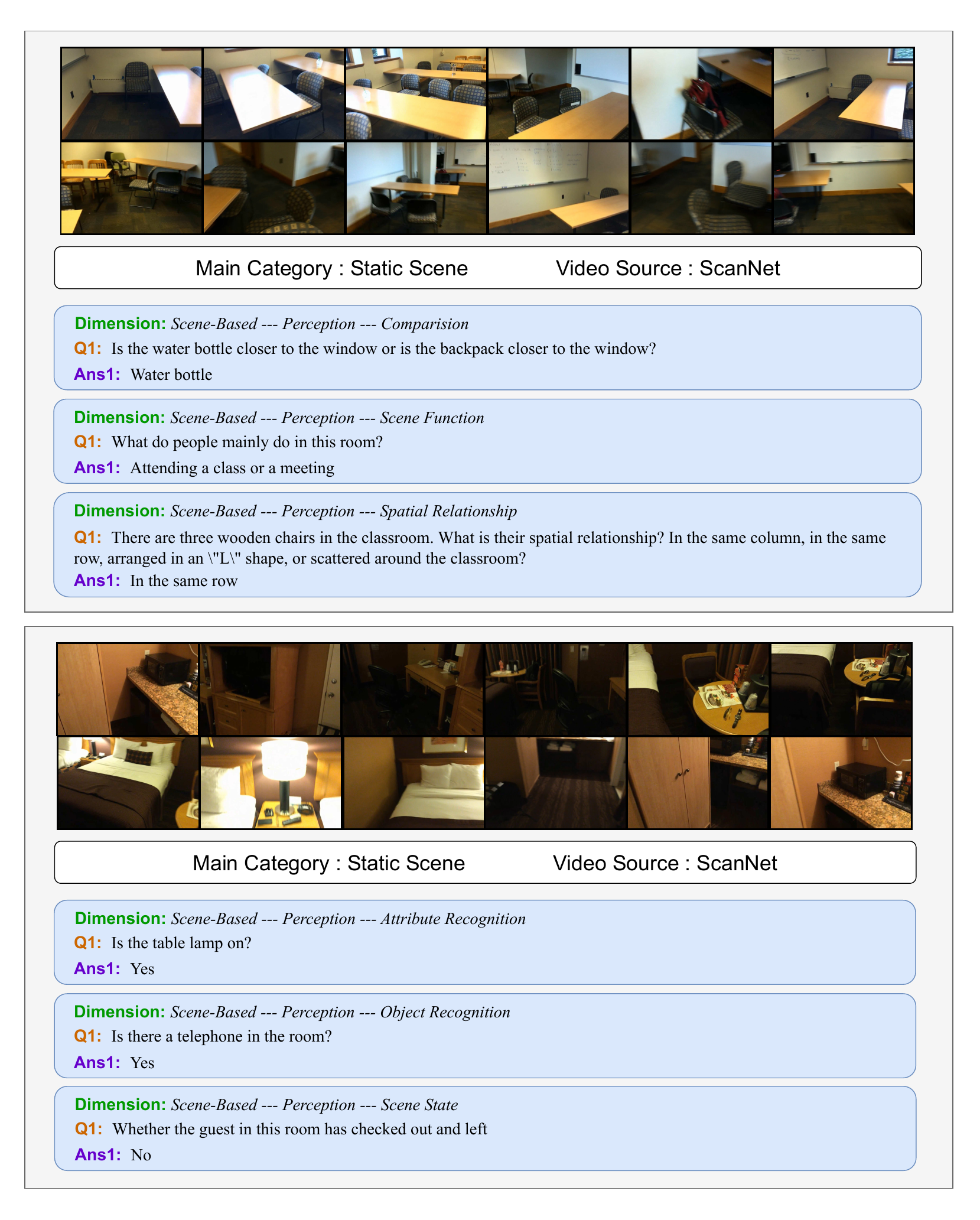}

   \caption{\textbf{Visualization of question answering examples in the static scene test set.} The main focus is on the scene-based category. Part 1 out of 2.}
   \label{fig:dimension_1}
\end{figure*}

\begin{figure*}[t]
  \centering
   \includegraphics[width=0.9\linewidth]{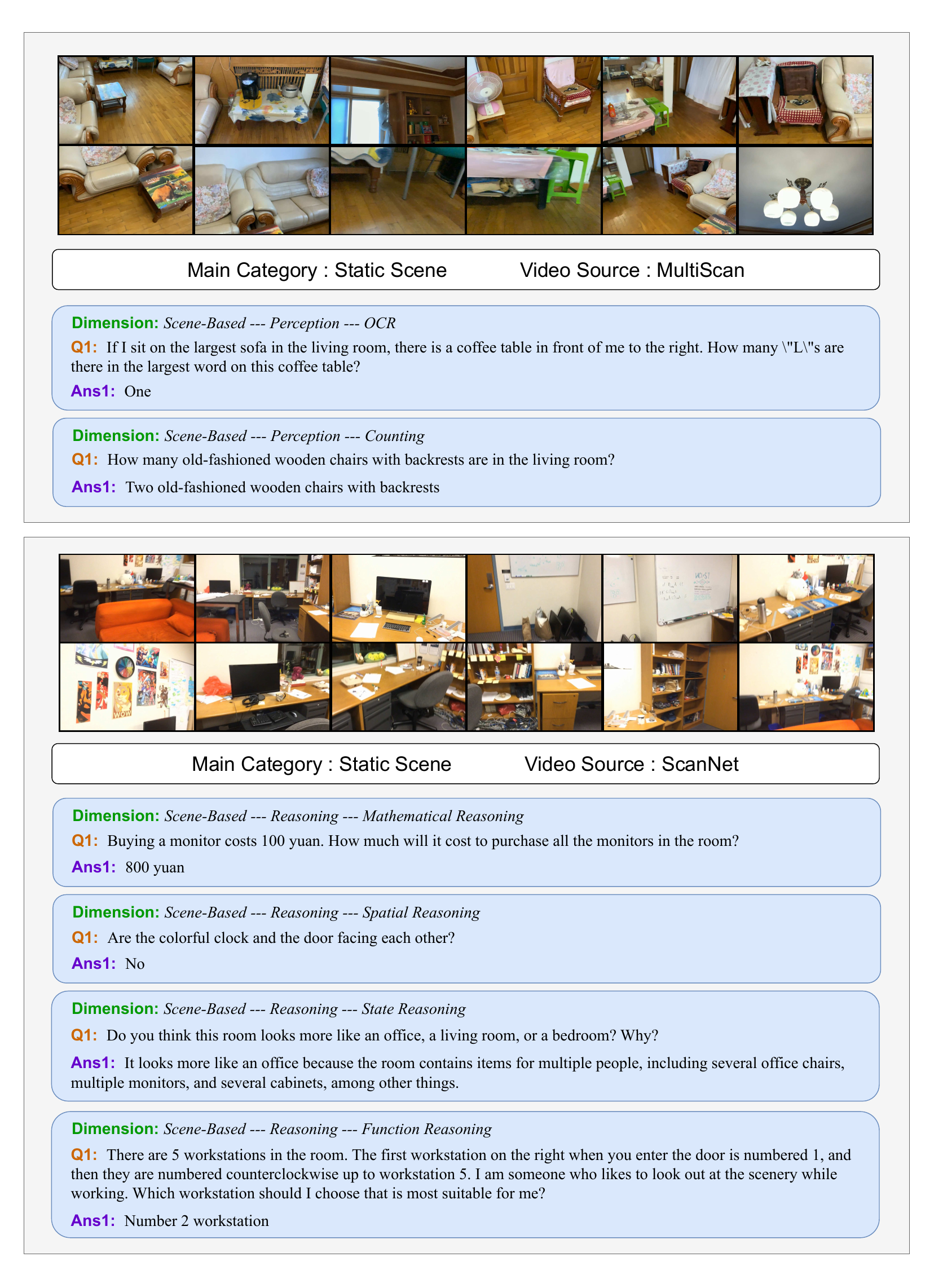}

   \caption{\textbf{Visualization of question answering examples in the static scene test set.} The main focus is on the scene-based category. Part 2 out of 2.}
   \label{fig:dimension_2}
\end{figure*}

Examples of all types of problems within the scene-based category can be found in Fig.~\ref{fig:dimension_1} and Fig.~\ref{fig:dimension_2}. We categorize scene-based questions into two types: perception and reasoning. Perception-type questions assess the capability of large vision-language models (LVLMs) to directly capture information from the environment, whereas reasoning-type questions require LVLMs to address more complex reasoning problems based on the environment information. Consequently, perception serves as the foundation for LVLMs in solving reasoning problems.

\paragraph{Perception}
\begin{itemize}
    \item \textbf{Comparison (Fig.~\ref{fig:dimension_1}):} Comparing certain information within a scene, such as the size of objects and their distances.
    \item  \textbf{Scene Function (Fig.~\ref{fig:dimension_1}):} The comprehension of the overall function of a scene, such as meetings, classes, or bathing.
    \item \textbf{Spatial Relationship (Fig.~\ref{fig:dimension_1}):} The understanding of spatial relationships between multiple objects, including concepts such as entangle, coverage, and verticality.
    \item \textbf{Attribute Recognition (Fig.~\ref{fig:dimension_1}):} The identification of attributes of individual objects, such as color, condition, and shape.
    \item  \textbf{Object Recognition (Fig.~\ref{fig:dimension_1}):} The acknowledgment of the existence and position of individual objects.
    \item  \textbf{Scene State (Fig.~\ref{fig:dimension_1}):} The identification of the overall state of a scene, including factors like tidiness, brightness, and clutter.
    \item \textbf{OCR (Fig.~\ref{fig:dimension_2}):} The recognition and analysis of textual content present within a scene.
    \item \textbf{Counting (Fig.~\ref{fig:dimension_2}):} To identify the quantity of certain objects in a scene or to comprehend reference resolution within a sequence.
\end{itemize}

\paragraph{Reasoning}

\begin{itemize}
    \item \textbf{Mathematical Reasoning (Fig.~\ref{fig:dimension_2}):} This involves possessing fundamental mathematical knowledge and utilizing it to address mathematical computation problems related to the physical world.
    \item \textbf{Spatial Reasoning (Fig.~\ref{fig:dimension_2}):} This requires an initial understanding of the spatial relationships between objects within a scene, followed by tackling more complex problems that arise from these spatial relationships.
    \item \textbf{State Reasoning  (Fig.~\ref{fig:dimension_2}):} This necessitates a foundational comprehension of the states of objects in a scene, followed by resolving more complex issues stemming from these states.
    \item \textbf{Function Reasoning (Fig.~\ref{fig:dimension_2}):} This entails first understanding the functions of objects within a scene, and subsequently addressing the more intricate problems that arise as a consequence of these functions.
\end{itemize}

\subsubsection{Robot-Centric}

\begin{figure*}[t]
  \centering
   \includegraphics[width=0.9\linewidth]{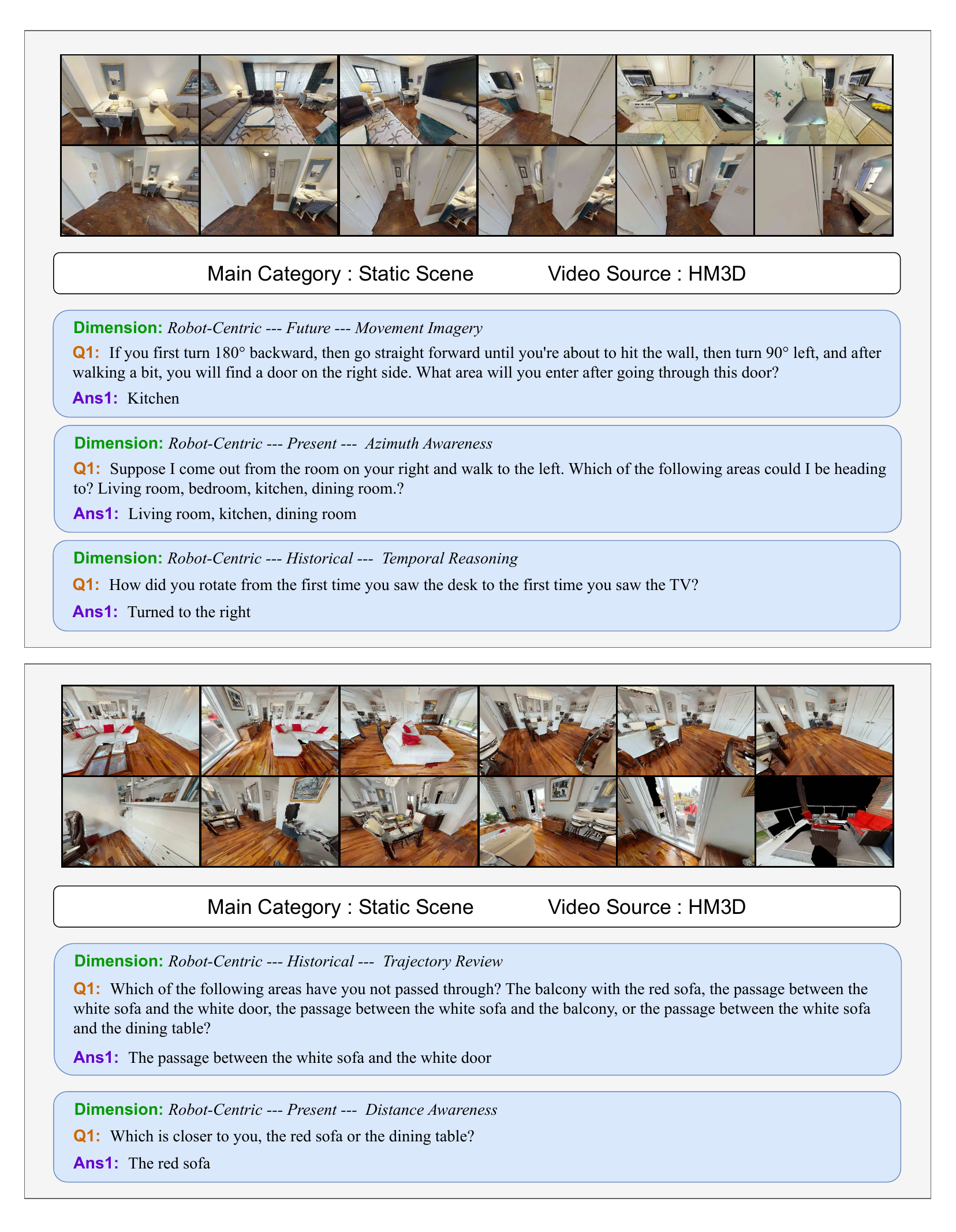}

   \caption{\textbf{Visualization of question answering examples in the static scene test set.} The main focus is on the robot-centric category. Part 1 out of 2.}
   \label{fig:dimension_3}
\end{figure*}

\begin{figure*}[t]
  \centering
   \includegraphics[width=0.9\linewidth]{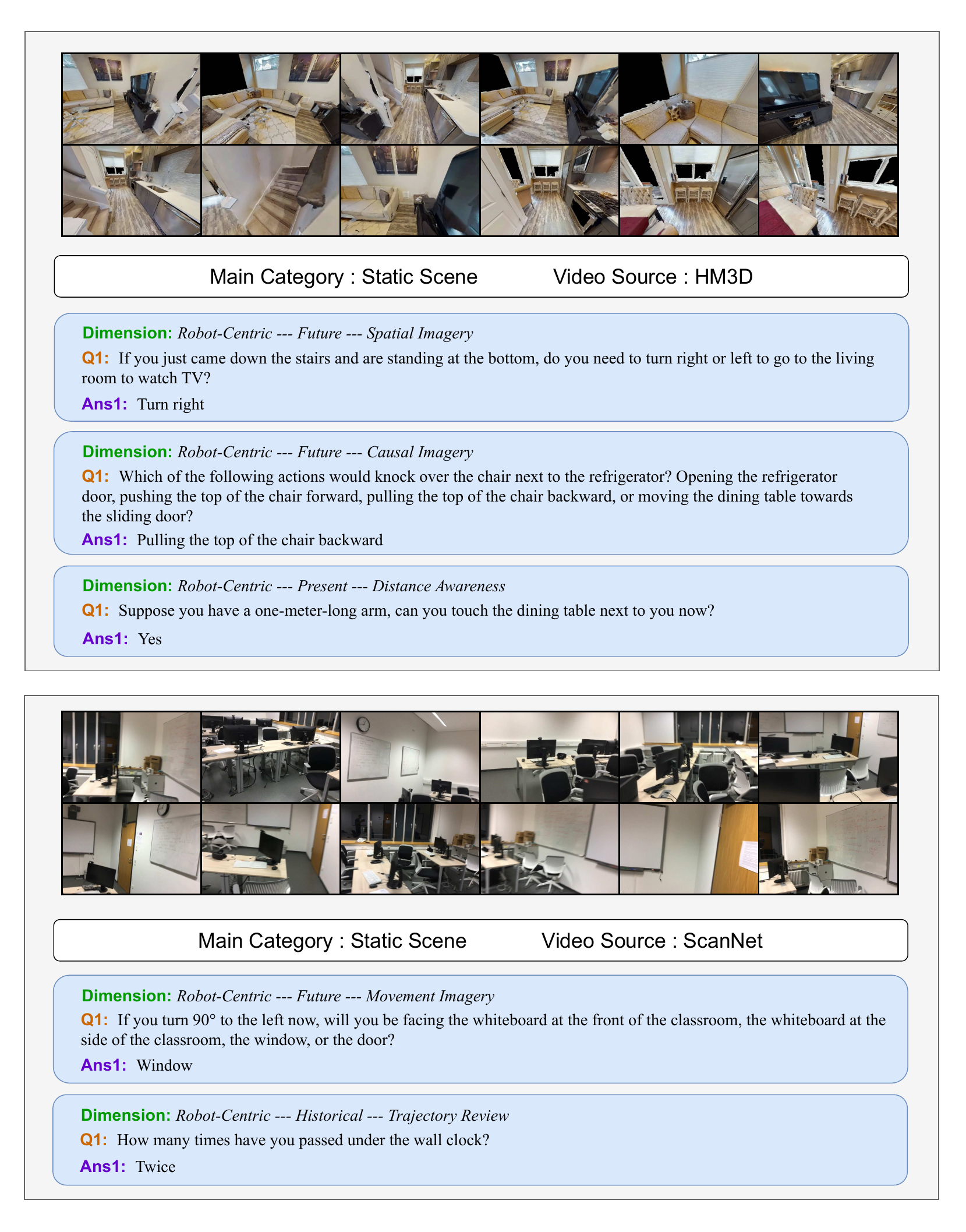}

   \caption{\textbf{Visualization of question answering examples in the static scene test set.} The main focus is on the robot-centric category. Part 2 out of 2.}
   \label{fig:dimension_4}
\end{figure*}

Examples of all types of questions within the robot-centric category can be found in Fig.~\ref{fig:dimension_3} and Fig.~\ref{fig:dimension_4}. It is important to note that robot-centric questions inherently assume the following fact: the subject of the video footage is a mobile robot. This robot is capable of moving forward and backward, rotating left and right, and tilting its head to observe upward and downward. All questions categorized as robot-centric pertain to the robot itself, rather than being solely answerable based on isolated scene information.
Based on the temporal dimensions of questions focus, we categorize robot-centric questions into three types: historical, present, and future.

\paragraph{Historical}
\begin{itemize}
    \item \textbf{Temporal Reasoning (Fig.~\ref{fig:dimension_3}):} Questions related to temporal elements such as time points, time intervals, and the sequencing of events based on the historical actions of the robot.
    \item \textbf{Trajectory Review (Fig.~\ref{fig:dimension_4}):} Retrospective examination of the robot's actions and path, addressing issues pertinent to the robot's spatial movement.
\end{itemize}

\paragraph{Present}
\begin{itemize} 
    \item \textbf{Azimuth Awareness (Fig.~\ref{fig:dimension_3}):} The understanding of the positional relationship between the robot and its environment based on the robot’s current location and orientation (as indicated by the last frame).
    \item \textbf{Distance Awareness (Fig.~\ref{fig:dimension_3}, Fig.~\ref{fig:dimension_4}): } The comprehension of the distance relationship between the robot and its environment, derived from the robot's current position and orientation (as indicated by the last frame).
\end{itemize}

\paragraph{Future}

\begin{itemize}
    \item \textbf{Movement Imagery (Fig.~\ref{fig:dimension_3}):} Envisioning the robot's future movements and the resultant changes in its spatial position.
    \item \textbf{Spatial Imagery (Fig.~\ref{fig:dimension_4}):} Imagining a hypothetical scenario in which the robot occupies a certain location, thereby analyzing the spatial relationships between the robot and its surrounding environment.
    \item \textbf{Causal Imagery (Fig.~\ref{fig:dimension_4}):} Imagining the potential impacts that certain actions taken by the robot would have on the environment.
\end{itemize}

\subsection{Dynamic Scene}

\begin{figure*}[t]
  \centering
   \includegraphics[width=0.9\linewidth]{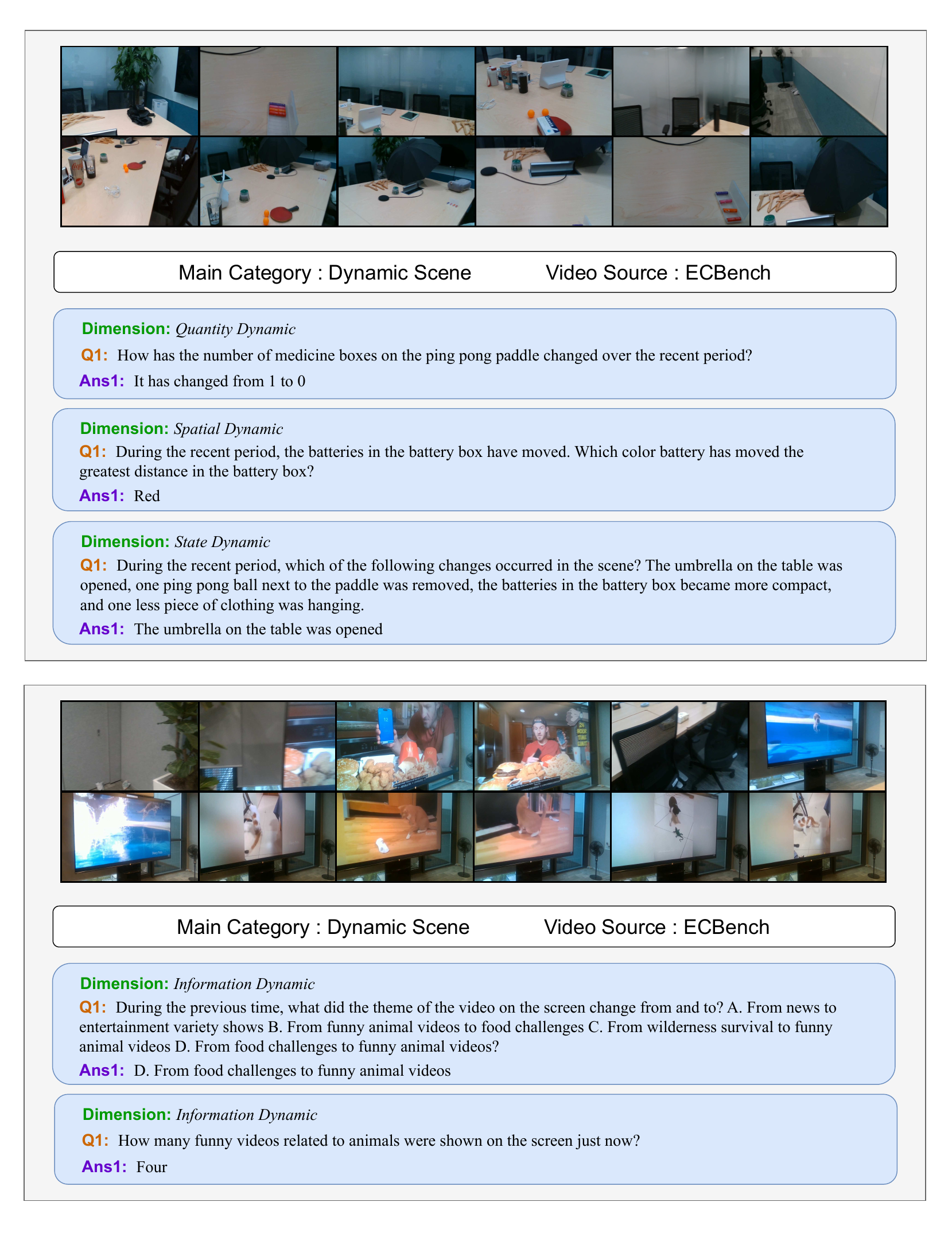}

   \caption{\textbf{Visualization of question answering examples in the dynamic scene test set.}}
   \label{fig:dimension_5}
\end{figure*}

\begin{itemize}
    \item \textbf{Quantity Dynamic (Fig.~\ref{fig:dimension_5}):} The increases, decreases, and transfers of the quantity of objects within an environment.
    \item \textbf{Spatial Dynamic (Fig.~\ref{fig:dimension_5}):} The changes in the spatial positions of individual objects, such as changes in direction, alterations in speed, and movement trajectories. The changes in the spatial relationships among multiple objects, such as parallelism and perpendicularity, degrees of overlap, and arrangements mode.
    \item  \textbf{State Dynamic (Fig.~\ref{fig:dimension_5}):} The state changes of objects, such as the on/off status of a light, the fullness or emptiness of a bottle, and the folding or unfolding of items. The state changes of scenes, such as brightness and darkness, as well as orderliness versus disorder.
    \item  \textbf{Information Dynamic (Fig.~\ref{fig:dimension_5}):} The dynamics of textual information on a two-dimensional plane, such as the content, color, and typography of the text. The dynamics in the content displayed on screens, such as video playback and content transitions.
\end{itemize}

\subsection{Hallucination}
\paragraph{Over-Confidence in Common Sense}

To investigate whether the current LVLMs exhibit hallucinations due to an excessive reliance on scene knowledge when handling embodied tasks, we specifically collect a set of counterfactual scenarios and pertinent questions to assess the LVLMs.

\begin{figure*}[t]
  \centering
   \includegraphics[width=0.9\linewidth]{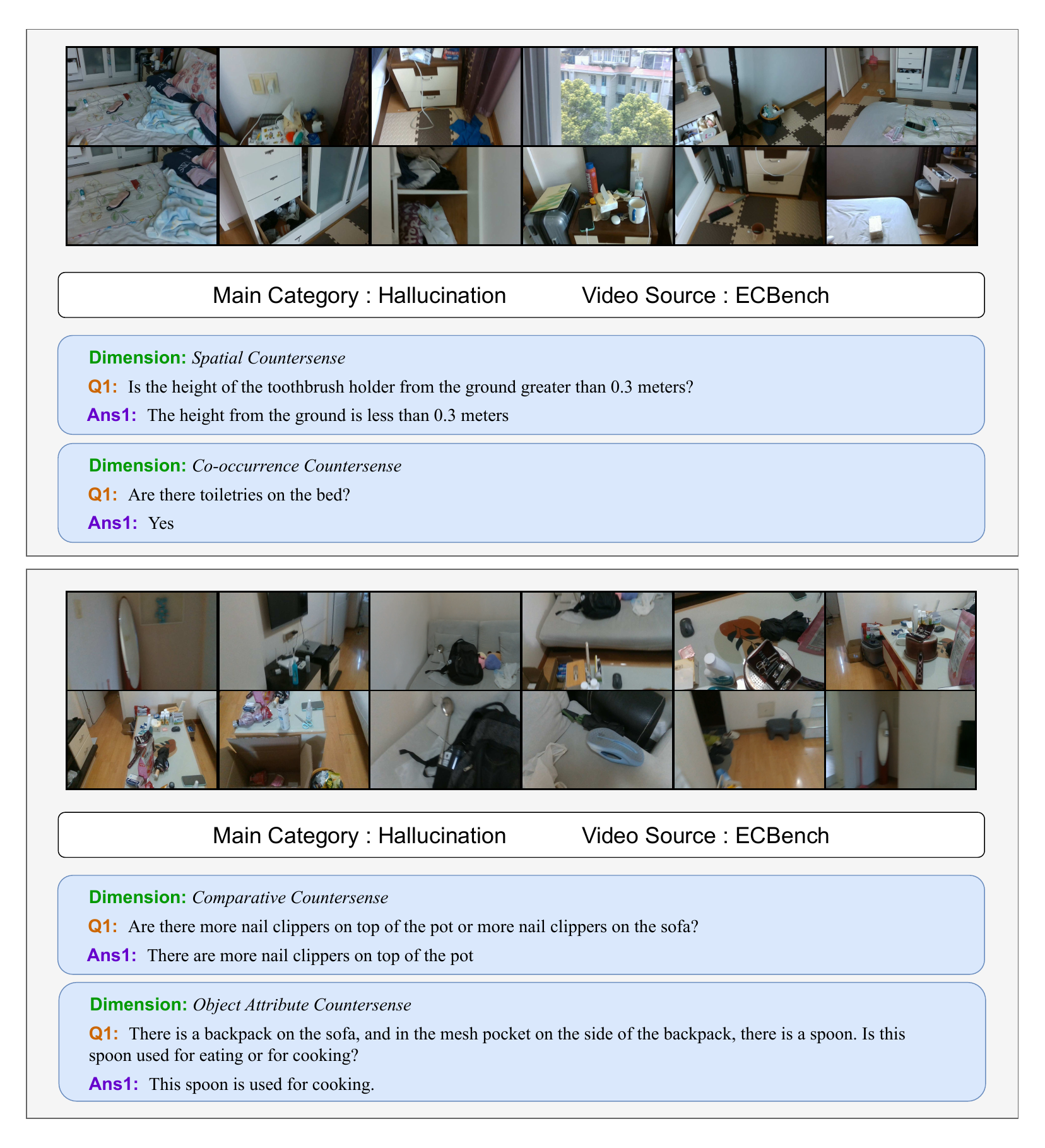}

   \caption{\textbf{Visualization of question answering examples in the hallucination test set.} The main focus is on the commensense over-confidence category.}
   \label{fig:dimension_6}
\end{figure*}

\newpage
\begin{itemize}
    \item \textbf{Spatial Countersense (Fig.~\ref{fig:dimension_6}):} 
    This refers to instances where spatial relationships do not conform to conventional expectations, such as a mouse pad placed atop a mouse, or a weighing scale situated on a dining table.

    \item \textbf{Object Attribute Countersense (Fig.~\ref{fig:dimension_6}):} This involves objects possessing attributes or characteristics that are relatively uncommon, for instance, a basketball smaller than a mobile phone, or a purple toilet.

    \item \textbf{Co-occurrence Countersense (Fig.~\ref{fig:dimension_6}):}  Unrelated objects are put together, and objects typically associated with one another are deliberately separated. For example, a pot and a comb do not frequently co-occur, whereas a toothbrush and a toothbrush holder commonly do.

    \item \textbf{Comparative Countersense (Fig.~\ref{fig:dimension_6}):} Unexpected outcomes in comparative assessments, such as a computer being closer to a toilet than a roll of toilet paper is, or a garbage bin being positioned at a height greater than that of a sink.

\end{itemize}

\paragraph{Over-Confidence in User Input}
In online dialogues, the trust that LVLMs place in user input is understandable. However, in physical world interactions mediated by robots, questioning user instructions and seeking clarification is fundamental to ensuring smooth human-robot interaction.

\begin{figure*}[t]
  \centering
   \includegraphics[width=0.9\linewidth]{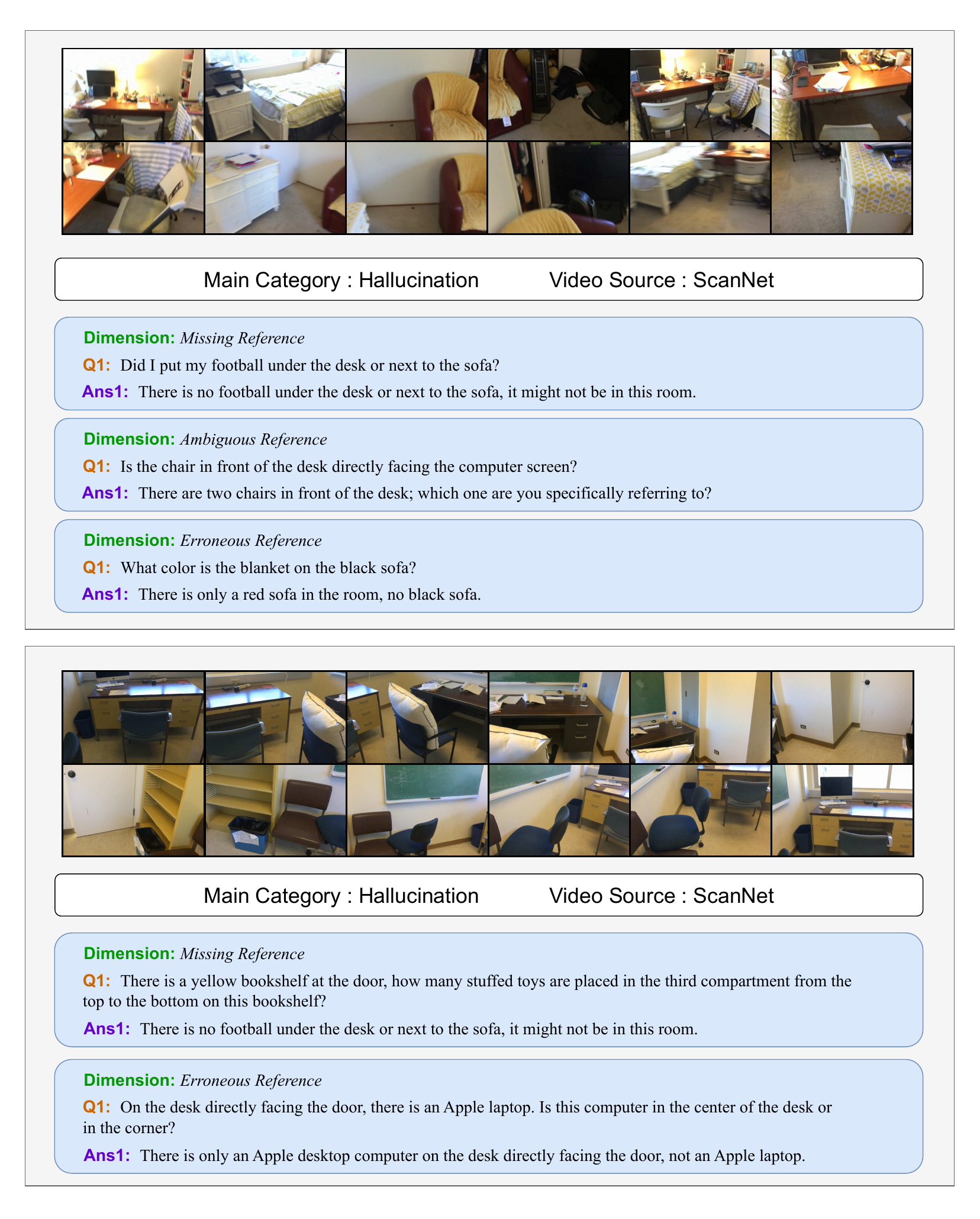}

   \caption{\textbf{Visualization of question answering examples in the hallucination test set.} The main focus is on the user-input over-confidence category.}
   \label{fig:dimension_7}
\end{figure*}

\begin{itemize}
    \item \textbf{Missing Reference (Fig.~\ref{fig:dimension_7}):} 
    When certain key referents in the user's instructions are absent from the scene, the model should identify these missing references.
    \item \textbf{Erroneous Reference (Fig.~\ref{fig:dimension_7}):}  
    When the user’s instructions contain erroneous references, the model should autonomously correct these errors and provide feedback to the user.
    \item \textbf{Ambiguous Reference (Fig.~\ref{fig:dimension_7}):} 
    When certain references in the user's instructions correspond to multiple entities in the scene, the model should enumerate these referents and inquire with the user for clarification on the intended reference.
\end{itemize}

\clearpage

\clearpage

\section{Construction of Evaluation Framework}
\label{sec:Evaluation Framework}
As mentioned in~\cref{sec:Benchmark Construction}, we design ECEval, a fine-grain evaluation framework that considers different types of tasks. That is, we also annotate a partially correct answer for each question in addition to a completely correct answer. Specifically, we first design some open-ended question-answer pairs as examples and then instruct GPT-4o to classify each question-answer into two types: closed-ended and open-ended. For each open-ended answer, we additionally manually annotate a 0.5-point answer as the reference. When scoring each model's responses, gpt-4o provides a continuous score between 0 and 1 based on the 1-point answer (100\% correct) and the 0.5-point answer (50\% correct). The prompt for LLM-assisted classification is as follows:
\lstset{
    framesep = 20pt,
    rulesep = 10pt,
    backgroundcolor = \color[RGB]{245,245,244},
    breaklines = true,
    breakindent = 0pt,
    basicstyle = \ttfamily\small,
    escapeinside = {(*@}{@*)} 
}

\begin{lstlisting}
(*@\color{red}{System Prompt}@*): You are a helpful assistant. You help me label questions that belong to free-response questions. Please note that if it is a multiple-choice question, a numerical question, counting, OCR, a true/false question, or a status judgment question or similar questions with definite answers, it does not belong to the free-response question type. If you determine that this question belongs to the free-response question type. Please output ###Free Response: True, otherwise ###Free Response: False.     
(*@\color{red}{USER}@*): 
###Question: If you pick up the footstool between the sofa and the TV and flip it 180 angle upside down, which objects in the scene will be affected and how? 
###Label Answer: Three handles and one remote control will fall onto the carpet.
(*@\color{red}{Assistant}@*): 
###Free Response: True 
(*@\color{red}{USER}@*): 
###Question: {(*@\color{blue}{example query}@*)} 
###Label Answer: {(*@\color{blue}{example answer}@*)}.
(*@\color{red}{Assistant}@*): 
###Free Response: {(*@\color{blue}{example Judgement}@*)}
....
(*@\color{red}{USER}@*): 
###Question: {(*@\color{blue}{user query}@*)} 
###Label Answer: {(*@\color{blue}{labed answer}@*)}. 
(*@\color{red}{Assistant}@*): 
\end{lstlisting}

\section{Additional Dataset Analysis}
\label{sec:Additional Dataset Analysis}

\begin{figure*}[t]
  \centering
  \begin{subfigure}{0.48\linewidth}
   \includegraphics[width=0.99\linewidth]{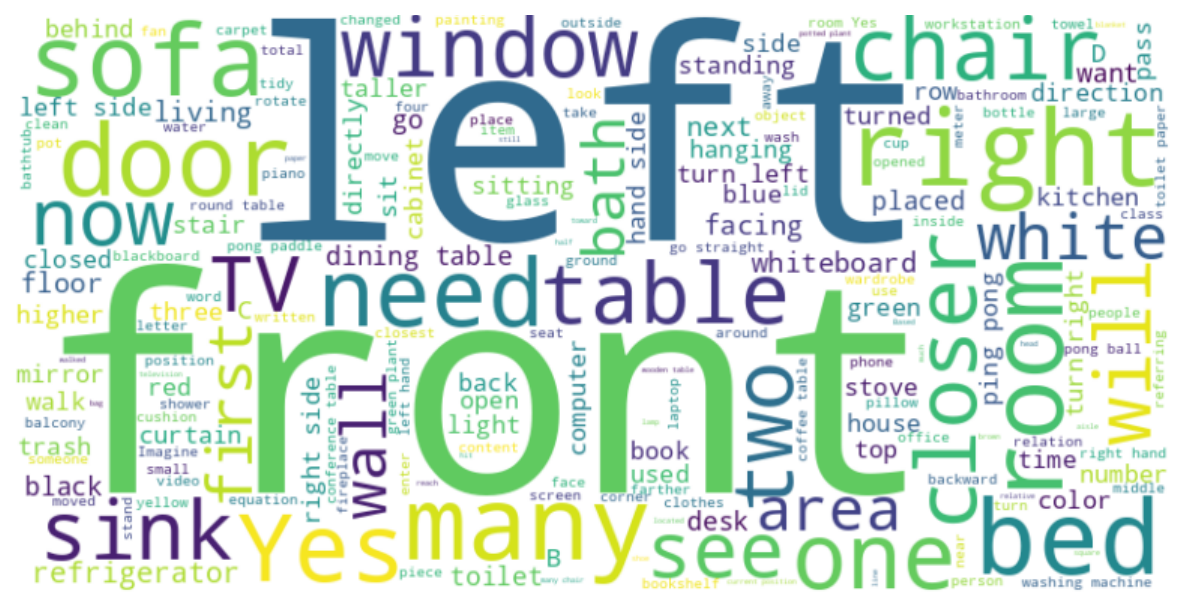}
   \caption{Word cloud of questions in ECBench.}
   \label{fig:wordcloud}
  \end{subfigure}
  \hfill
  \begin{subfigure}{0.48\linewidth}
    \includegraphics[width=0.99\linewidth]{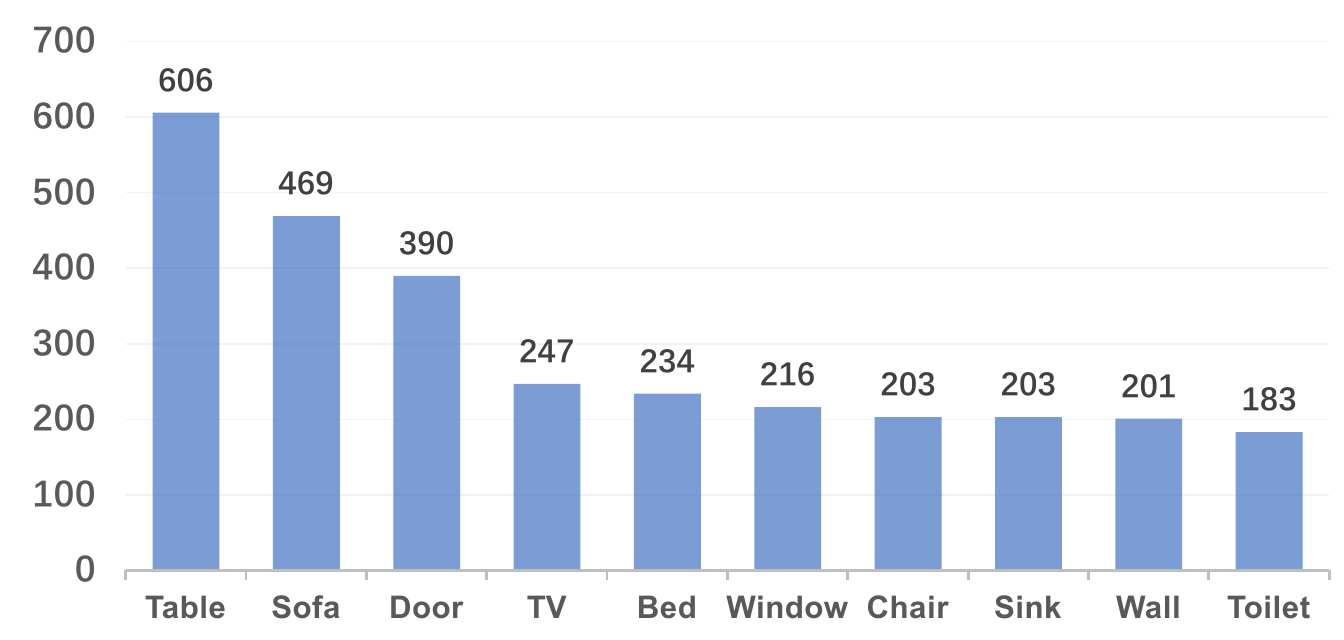}
   \caption{The top ten object nouns by word frequency in ECBench.}
   \label{fig:word_count}
  \end{subfigure}
  \caption{A more detailed analysis of the textual content of ECBench.}
  \label{fig:more word analysis}
\end{figure*}

In Sec.~\ref{sec: Dataset Statistics} of the main text, we conduct a comprehensive quantitative analysis of ECBench. This section delves into a more detailed examination of the question-answer text within ECBench. In Fig.~\ref{fig:wordcloud}, we present a word cloud depicting all the questions in ECBench. The most frequently occurring terms are predominantly spatial pronouns, such as "left" and "front." These words are crucial for robot-centric referentiality and are among the most commonly used vocabulary in our interactions with robots. Additionally, ECBench contains numerous object nouns, including "sofa" and "chair," which represent the most common items within indoor environments and are frequently mentioned by users.

In Fig.~\ref{fig:word_count}, we enumerate the top ten most prevalent object nouns in ECBench, with "table" appearing most frequently at 606 occurrences. Overall, ECBench encompasses 1,067 distinct object nouns, exhibiting a richness in object referentiality that far surpasses that of other embodied question-answering datasets \cite{openeqa, scanqa}.

\section{More Details of Experiment}
For all experiments, we set the temperature to 0.2. For the GPT family models, we request the endpoint of GPT-4o-0513 and GPT-4o-mini-0513 through the Azure\footnote{\url{https://azure.microsoft.com/en-us/}}. For Qwen2VL-72B/7B and InternVL2, we use vllm\footnote{\url{https://docs.vllm.ai/en/latest/}} for employment. To score each model's prediction, we uniformly employ GPT-4o-0513 to evaluate the predicted answers based on labeled answer and 0.5-point answer. We employ the same prompt for benchmarking and evaluating stages across all experiments as follows:
\lstset{
    framesep = 20pt,
    rulesep = 10pt,
    backgroundcolor = \color[RGB]{245,245,244},
    breaklines = true,
    breakindent = 0pt,
    basicstyle = \ttfamily\small,
    escapeinside = {(*@}{@*)} 
}

\begin{lstlisting}
(*@\textbf{Benchmarking Each Model}@*)
(*@\color{red}{System Prompt}@*): You are moving in an indoor environment. The image sequence is the scene you just saw. You are now staying at the last frame of the video. Please answer the question with one word or one sentence, as concise and accurate as possible.     
(*@\color{red}{USER}@*): The question: {(*@\color{blue}{user query}@*)}

(*@\textbf{Scoring Open-ended Question}@*)
(*@\color{red}{System Prompt}@*): You are a helpful assistant.  
Please score the predicted answer according to the given question and huamn labeled the 5-score answer, and the 3-score answer. 0 score represents completely wrong, 5 scores represents completely correct, and 3 scores represents partially correct. Please refer to them to score the predicted answer: [0, 1, 2, 3, 4, 5]. You need to consider the answer from two perspectives: accuracy and completeness. Output ###Judge:
(*@\color{red}{USER}@*): 
###Question: If you pick up the footstool between the sofa and the TV and flip it 180 degree upside down, which objects in the scene will be affected and how? 
###5 Score Answer: Three handles and one remote control will fall onto the carpet. 
###3 Score answer: The remote control will fall onto the carpet. 
###Predicted Answer: Remote control will fall onto the carpet.
(*@\color{red}{Assistant}@*): ###Judge: 3
(*@\color{red}{USER}@*): 
###Question: If you pick up the footstool between the sofa and the TV and flip it 180 degree upside down, which objects in the scene will be affected and how? 
###5 Score Answer: Three handles and one remote control will fall onto the carpet. 
###3 Score answer: The remote control will fall onto the carpet. 
###Predicted Answer: handles.
(*@\color{red}{Assistant}@*): ###Judge: 2
...
{(*@\color{blue}{Another 3 Examples}@*)}
...
(*@\color{red}{USER}@*): 
###Question: {(*@\color{blue}{user query}@*)}
###5 Score Answer: {(*@\color{blue}{5 Score Answer}@*)}
###3 Score Answer: {(*@\color{blue}{3 Score Answer}@*)}
###Predicted Answer: {(*@\color{blue}{Predicted Answer}@*)}
(*@\color{red}{Assistant}@*):

(*@\textbf{Scoring Closed-ended Question}@*)
(*@\color{red}{System Prompt}@*): You are a helpful assistant. Please judge whether the predicted answer is correct or not according to the given question and labeled answer. You need to consider the answer from two perspectives: accuracy and completeness. Output ###Judge: True only when the predicted answer is accurate and complete; otherwise, output ###Judge: False
(*@\color{red}{USER}@*): 
###Question: If you are now standing outside the restroom stall door away from the sink. Which edge will the door rotate around if you move forward? 
###Label Answer: Using the left side edge as the axis. 
###Predicted Answer: The door will rotate around the left edge
(*@\color{red}{Assistant}@*): ###Judge: True
...
{(*@\color{blue}{Another 4 Examples}@*)}
...
(*@\color{red}{USER}@*): 
###Question: {(*@\color{blue}{user query}@*)}
###Label Answer: {(*@\color{blue}{label answer}@*)}
###Predicted Answer: {(*@\color{blue}{Predicted answer}@*)}
(*@\color{red}{Assistant}@*):
\end{lstlisting}

\label{sec:More Details of Experiment}

\section{Limitations and Broader Impacts}
\paragraph{Limitations}

In this study, we introduce ECBench, a novel open-world embodied cognition benchmark. ECBench represents the first effort to systematically analyze embodied cognitive issues and to establish a comprehensive evaluation framework. However, several limitations warrant further investigation by future researchers:

\begin{enumerate}[label=\arabic*., align=left, left=0pt]
    \item Due to budget constraints, we are unable to test all proprietary models, such as Claude-3.5, among others.
    \item Theoretically, robots may possess a longer video memory during the execution of specific tasks; however, the video inputs for ECBench are limited to under five minutes, which may not adequately evaluate the cognitive abilities of LVLMs regarding prolonged visual memory.
    \item Cognition of dynamic scenarios in open-world environments is a previously unexplored area, leading to a lack of diverse dynamic scene videos. Although we make efforts to collect some real-world dynamic scene videos, these still fall short in terms of scene richness and object variety.
    \item The natural interaction between robots and humans should not be a singular question-and-answer exchange, but rather a streaming and interwoven dialogue. Therefore, the evolution of ECBench towards a more practical and flexible question-and-answer format will be a key direction for our future research.
\end{enumerate}

\paragraph{Broader Impacts}

As an evaluation benchmark focused on the domain of embodiment, ECBench is poised to attract significant attention from researchers interested in exploring the cognitive processes of perceiving the real world through RGB-D videos. Furthermore, ECBench aims to facilitate the current LVLMs in transcending the limitations posed by online images and videos, thereby shifting their focus more towards the visual input modalities that robots encounter in the real world. The development of world models based on RGB-D video is a shared aspiration among AI researchers. We also hope that ECBench will contribute substantially to the advancement of robotic visual cognition capabilities.